\documentclass{article} 
\usepackage{iclr2022_conference,times}

\usepackage{amsmath,amsfonts,bm}

\def\eqref#1{equation~\ref{#1}}
\def\Eqref#1{Equation~\ref{#1}}

\def\1{\bm{1}}

\DeclareMathAlphabet{\mathsfit}{\encodingdefault}{\sfdefault}{m}{sl}
\SetMathAlphabet{\mathsfit}{bold}{\encodingdefault}{\sfdefault}{bx}{n}

\newcommand{\E}{\mathbb{E}}

\newcommand{\R}{\mathbb{R}}

\newcommand{\Var}{\mathrm{Var}}

\DeclareMathOperator*{\argmax}{arg\,max}

\usepackage{hyperref}
\usepackage{url}
\usepackage{algorithm}
\usepackage{tabularx}
\usepackage{graphicx}
\usepackage{tikz}
\usepackage{pdflscape}
\usepackage{xspace}
\usepackage{subcaption}
\usepackage{wrapfig}
\usepackage{amsmath,amsthm,amssymb}
\usepackage[ruled,vlined,algo2e]{algorithm2e}
\newcommand*{\defeq}{\stackrel{\text{def}}{=}}
\DeclareMathOperator*{\mmax}{mellowmax}
\DeclareMathOperator*{\wmax}{mmax}
\newtheorem{theorem}{Theorem}[section]

\newtheorem{lemma}[theorem]{Lemma}

\newcommand*{\C}[1]{\mathcal{#1}}
\usepackage{color}

\usepackage{authblk}

\title{Temporal-Difference Value Estimation via Uncertainty-Guided Soft Updates}
\iclrfinalcopy

\author[1,\thanks{Correspondence to: litianl1@uci.edu}]{Litian Liang}
\author[1]{\ Yaosheng Xu}
\author[1]{\ Stephen McAleer}
\author[1]{\ Dailin Hu}
\author[1]{\\Alexander Ihler}
\author[2]{\ Pieter Abbeel}
\author[1]{\ Roy Fox}
\affil[1]{Department of Computer Science, University of California, Irvine}
\affil[2]{Department of Electrical Engineering and Computer Science, \linebreak University of California, Berkeley}

\begin{document}

\maketitle

\begin{abstract}
Temporal-Difference (TD) learning methods, such as Q-Learning, have proven effective at learning a policy to perform control tasks. One issue with methods like Q-Learning is that the value update introduces bias when predicting the TD target of a unfamiliar state. Estimation noise becomes a bias after the max operator in the policy improvement step, and carries over to value estimations of other states, causing Q-Learning to overestimate the Q value. Algorithms like Soft Q-Learning (SQL) introduce the notion of a soft-greedy policy, which reduces the estimation bias via soft updates in early stages of training. However, the inverse temperature $\beta$ that controls the softness of an update is usually set by a hand-designed heuristic, which can be inaccurate at capturing the uncertainty in the target estimate. Under the belief that $\beta$ is closely related to the (state dependent) model uncertainty, Entropy Regularized Q-Learning (EQL) further introduces a principled scheduling of $\beta$ by maintaining a collection of the model parameters that characterizes model uncertainty. In this paper, we present Unbiased Soft Q-Learning (UQL), which extends the work of EQL from two action, finite state spaces to multi-action, infinite state space Markov Decision Processes. We also provide a principled numerical scheduling of $\beta$, extended from SQL and using model uncertainty, during the optimization process. We show the theoretical guarantees and the effectiveness of this update method in experiments on several discrete control environments.
\end{abstract}

\section{Introduction}
Reinforcement Learning (RL) algorithms learn a control policy that maximizes the expected discounted sum of future rewards (the \emph{policy value}) from state, action, and reward experience collected through interactions with an environment. 
Temporal-Difference (TD) RL methods \citep{sutton2018reinforcement} maintain value estimates, and iteratively improve them by combining experienced short-term rewards with estimates of the following long-term values. Q-Learning \citep{watkins1992q} is a popular TD method that learns state–action value estimates (\emph{Q function}), by assuming for target value estimates that a greedy policy will be used in the following state. 

Although Q-learning asymptotically converges to the optimal policy, it is known to be positively biased and overestimate the value before convergence \citep{Thrun-1993-15908}. This bias is detrimental to efficient learning, because it can cause optimal actions to appear suboptimal, delaying the collection of experience relevant to the agent's optimal policy. Several approaches have been proposed to mitigate the overestimation in TD learning, including Double Q-Learning \citep{DBLP:journals/corr/HasseltGS15}, which attempts to correct this bias by using two Q functions; Soft Q-Learning (SQL)~\citep{Fox2016Taming, DBLP:journals/corr/HaarnojaTAL17}, in which value updates use a softmax policy for target value estimation; and Maxmin~\citep{lan2020maxmin}, which estimates the target value as the minimum of several Q functions. None of these methods eliminate the bias completely. On the other hand, it should be noted that the role that unbiased estimation could play in improving reinforcement learning remains unclear~\citep{lan2020maxmin}.



\citet{Fox2016Taming} showed that in SQL there exists a particular state-dependant inverse-temperature parameter $\beta$ of the softmax operator that makes the target value estimates unbiased. However, this unbiased-update $\beta$ value is generally unknown. 
\citet{Fox2019Toward} 
suggests that $\beta$ is inversely related to the variance of the estimate's parameter distribution by deriving a closed-form expression for MDPs with two actions. However, it is not obvious how to generalize this expression to multi-action Markov Decision Processes and neural network $Q$ function approximator representations.

In this work, we  estimate $\beta$ by maintaining an ensemble of Q functions 
to approximate the uncertainty in Q.
This estimated $\beta$ provides a principled temperature schedule that is ``softer'' when uncertainty is high but gradually becomes ``harder'' as training proceeds. 
We experiment in 
Atari environments to show the effectiveness of our method in practical discrete action space environments, and find that our method outperforms the Rainbow DQN \citep{hessel2018rainbow} and PPO \citep{schulman2017proximal} algorithms in most Atari environments. This work shows that incorporating awareness of model uncertainty can help to provide a target with less bias, which leads to faster policy improvement. 

To summarize, our contributions are as follows:
\begin{itemize}
    \item We provide a principled numerical method for estimating a $\beta$ term that makes the maximum-entropy target value approximately unbiased in SQL.
    \item We provide extensive experimental results demonstrating that reducing bias in SQL via our estimated $\beta$ improves performance.
    \item We provide a proof of convergence of our method. 
\end{itemize}


\section{Preliminaries}
We consider a Markov Decision Process (MDP) with transition probability $p(s'|s,a)$. An agent controls the process using policy $\pi(a|s)$. A reward $r$ that only depends on $s$ and $a$ is observed after an action is taken. This work focuses on MDPs with discrete action spaces, $a \in \{0, 1, ... d - 1\}$, where $d$ is the size of the discrete action space and varies in different environments.

We denote trajectory as a sequence of states, actions, and rewards $\xi = (s_0, a_0, r_0, s_1 \ldots)$. An RL algorithm should discover a control policy that maximizes expected return of trajectory $R(\xi) = \sum_{t\ge0} \gamma^t r_t$, where $t$ is the time step in an interaction process and $0\le\gamma\le1$ is a discount factor. Value-based methods infer a control policy by maintaining a state-action value function of some policy $\pi$,
$$Q_{\pi}(s,a) = \E_{a\sim\pi(\cdot|s)}[R(\xi)|s_0=s, a_0=a].$$
Q-Learning \citep{watkins1992q} 
learns the optimal value of each state action by minimizing the Temporal-Difference loss via the stochastic update,
$$Q(s,a) \leftarrow Q(s,a) + \alpha(r + \gamma \max_{a'}Q(s',a') - Q(s,a)),$$
This update is guaranteed to converge to the optimal $Q^*$, since the Bellman operator for Q-Learning,
$$\C{B}[Q](s,a) = \E[r(s,a)] + \gamma  \E_{s'\sim p(\cdot|s,a)}\big[\max_{a'}Q(s',a')\big]$$
is a contraction in $L^{\infty}$.

However, in early stages of training, our estimate $Q$ can have significant errors (``noise''). 
By using this noisy estimate, Q-Learning results in biased estimation of the next state value $V(s)=\max_a Q(s,a)$ \citep{Thrun-1993-15908}. The Q-Learning target estimation bias is a consequence of Jensen's inequality, since
$$\E_Q\big[\max_a Q(s,a)\big] \geq \max_a\E_Q\big[ Q(s,a) \big].$$
Training with this target estimate will propagate the bias on to preceding states due to the nature of the Bellman backup \citep{NEURIPS2019_c2073ffa, NEURIPS2020_d7f426cc}. 
Deep Q-Networks \citep{DBLP:journals/corr/MnihKSGAWR13,mnih2015human} 
effectively incorporate the same biased estimate, training a neural network on squared error between the current $Q$ and the Bellman target:
$$\mathcal{L}(s,a,r,s',\theta)=(r+ \gamma \max_{a'} Q_{\bar{\theta}}(s',a')-Q_\theta(s,a))^2 $$

Instead, we would prefer an estimator of $Q$ with an unbiased estimate of the optimal value:
$$\E \big[ V_{\text{unbiased}}(s) \big] = \max_a \E_Q\big[ Q(s,a) \big]$$
A greedy policy $a=\argmax_a Q(s,a)$ selects the best action $a$ given its current knowledge of $Q(s,a)$ as the target estimate for the Bellman backup. This is problematic because in early stages of learning, the $Q(s,a)$ estimates are incorrect. \citet{Fox2016Taming} proposed an additional cost term, which combined the expected return induced by the policy with a penalty for deviating from an agnostic prior policy $\pi_0$, i.e.,
$$\Tilde{r_t} = r_t - \frac{1}{\beta}\log\frac{\pi(a|s)}{\pi_0(a|s)}$$
The state-action free energy function \citep{Fox2016Taming} evaluated at the soft-optimal policy can be written as $V_{\beta}(s)$, the mellowmax \citep{pmlr-v70-asadi17a} over the action value of the next state. 
$$V_\beta(s) = \frac{1}{\beta}\log\E_{a\sim\pi_0(\cdot|s)}\big[\exp(\beta Q(s,a))\big],$$

$V_\beta(s)$ is the soft-optimal value function under maximum-entropy notion of optimality, where $\beta$ is the inverse temperature.
\citet{Fox2016Taming} also proved that under a specific scheduling of $\beta$, controlling the level of trade-off between a deterministic policy ($\beta\rightarrow \infty$) and an agnostic prior policy ($\beta=0$) for selecting the distribution over actions to evaluate the next state, the soft-optimal value $V_\beta(s)$ of the maximum-entropy objective, is an unbiased estimator of the next state value.
The soft-greedy policy is the policy $\pi(a|s)$ that maximizes $V_\beta(s)$,
$$\pi(a|s) = \frac{\pi_0(a|s)\exp(\beta Q(s,a))}{\sum_{\bar{a}} \pi_0(a|s)\exp(\beta Q(s,\bar{a}))}.$$
Selecting the action soft-greedily using a good $\beta$ that captures the uncertainty in current estimates mitigates the positive bias caused by overestimating the action value. Previous work has also shown that selecting a good schedule of $\beta$ for balancing uncertainty and optimality is crucial for the success of maximum-entropy RL algorithms \citep{Fox2016Taming, haarnoja2018soft_app, Fox2019Toward, countbased2021, TESSAC2021}. We set to find the inverse temperature $\beta$ using model uncertainty.

\section{Why Reduce Estimation Bias}
RL with unbiased value estimation is not necessarily the best way to do exploration \citep{Fox2016Taming, lan2020maxmin}, but it is important to differentiate between exploration bonus and biased estimation. ``Optimism under uncertainty'' refers to the exploration bonus in selecting an action when that state-action pair is not familiar to the agent.  In contrast, bias corresponds to systematic errors in the value estimates themselves.  Since there is no reason to believe that this bias is somehow well-chosen to balance exploration and exploitation, it makes sense to attempt to decouple these two issues by performing unbiased estimation; then exploration bonuses can be incorporated in a more coherent way.

The TD estimation bias may also arise from multiple sources, some of which may be outside our control. The bias we attempt to eliminate in this work is that induced by the $\max$ operator in the TD next-state value estimate. If the agent is insufficiently familiar with the the next state and actions, the estimate can be noisy or wrong. For example, our initial Q values can be biased by initialization. If we have never encountered a state, it's value estimate is determined by generalization, which is known to introduce optimistic bias \citep{Thrun-1993-15908}. Even if we can estimate these values in a less biased way, our estimate errors will combine with the randomness inherent in the MDP, and create a large Jensen's gap between the Q target estimate and the true state value as we propagate our estimates through the backward recursive structure of the Bellman updates \citep{DBLP:journals/corr/abs-2007-04938, NEURIPS2019_c2073ffa, NEURIPS2020_d7f426cc}. This motivates us to examine the connection between our update policy and our uncertainty in Q.

\section{Unbiased Value Estimation}

\subsection{Relationship between Model Uncertainty and Policy Entropy}
Making decisions and learning with imperfect decision makers is well studied, e.g., \citet{rubin2012trading}. Ideally, we want a deterministic policy and ``hard max'' update only when the model is certain of its target estimate given the observed data, and the update and policy should be more stochastic when the target estimate is more uncertain. In maximum-entropy RL frameworks \citep{ziebart2010modeling}, the policy entropy that determines the greediness of each update is controlled by an inverse temperature $\beta$, analogously to the thermodynamic temperature of a system in a Boltzmann distribution. Unlike many maximum-entropy RL methods that use a constant temperature or a heuristic temperature schedule, we propose to estimate the $\beta(s)$ that yields an unbiased update during training.  Then, we select the next state action soft-greedily and perform a soft-optimal target update that is unbiased in expectation as in SQL \citep{Fox2016Taming}:
\begin{equation*}
Q(s,a) \leftarrow Q(s,a) + \alpha(r + \frac{\gamma}{\beta}\log\E_{a\sim\pi_0(a|s)}\big[\exp(\beta Q(s,a))\big] - Q(s,a)).
\end{equation*}


\subsection{Unbiased Soft Update via Model Uncertainty}
%
%
%

\cite{Fox2016Taming} showed the existence of a state-conditioned inverse temperature $\beta_Q(s)$, 
for which the state value function $V_\beta(s)$ is an unbiased estimator of $\max_a \E[Q(s,a)]$. 
Specifically, assuming our current estimate $Q$ is unbiased, there exists a temperature $\beta(s)$ whose maximum-entropy policy satisfies,
\begin{equation*}
    \E \big[ V_{\beta}(s) \big] = \E_Q \bigg[ \frac{1}{\beta_Q(s)}\log \E_{a\sim\pi_0(a|s)}\big[\exp(\beta_Q(s) Q(s,a))\big] \bigg] = \max_a \E_Q \bigg[ Q(s,a) \bigg]
\end{equation*}
and so does not accumulate value estimation bias.

In practice, we propose to estimate these expectations using a finite collection of functions $\{Q^{(i)}\}$; by defining the discrepancy function
\begin{equation*}
    f_{s;Q}(\beta) = \hat \E_Q\bigg[\frac{1}{\beta}\log \E_{a\sim\pi_0(a|s)}\big[\exp(\beta Q(s,a))\big]\bigg] - \max_a \hat \E_Q\bigg[ Q(s, a) \bigg],
\end{equation*}
where $\hat{\E}$ denotes the empirical average over the finite collection $\{Q^{(i)}\}$,
we can solve for the root of $f$ numerically to estimate $\beta(s)$.
The resulting $\beta$ depends on the model uncertainty --- if the collection $\{Q^{(i)}\}$ disagree on which action should achieve the maximum value (or the greedy policy), $\E_Q[V_\beta(s)] - \max_a \E_Q[Q(s,a)]$ can be large, resulting in $\beta(s)$ close to 0.
%

In practice, approximating the desired quantity $\max_a \E_Q[Q(s,a)]$ with our empirical average $\max_a \hat{\E}[Q(s,a)]$
still creates a slight positive bias, since by Jensen's inequality,
$$\max_a \E_Q\Big[ Q(s, a) \Big] \leq \E_{\{Q\}}\Big[\max_a \hat{\E}_Q\big[ Q(s, a) \big]\Big]. $$
To compensate, we correct the inverse temperature by multiplying by some $\kappa \in (0,1)$.

\section{Learning in Finite State Space MDPs}
We first consider finite state MDPs, in which $Q(s,a)$ is represented in a tabular form.
We give convergence properties of our ensemble approach, and illustrate its behavior on an example domain.


\subsection{Unbiased Soft Update}
In a tabular representation, given an ensemble $\{Q^{(i)}\}$, 
the update rule of Unbiased Soft Q-Learning (UQL) on each ensemble member $Q^{(i)}$ using experience $(s,a,r,s')$ is given by,
\begin{equation} \label{UQL_update}
Q^{(i)}(s,a) \leftarrow (1 - \alpha_t)Q^{(i)}(s,a) + \alpha_t \Big( r(s,a) + \frac{\gamma}{\kappa\beta_Q(s)}\log \E_{a\sim\pi_0(\cdot|s')}\big[\exp(\kappa\beta_Q(s) Q^{(i)}(s',a'))\big] \Big).
\end{equation}
We can define the Bellman operator at temperature $w = \frac{1}{\kappa\beta_Q(s)}$ by,
\begin{equation*}
    \C{B}_w[Q^{(i)}](s, a) \defeq \E[r(s,a)] +  \gamma \E_{s' | s, a \sim p} \Big[ w \log \E_{a\sim\pi_0(\cdot|s')}\big[\exp(Q^{(i)}(s',a') / w)\big] \Big].
\end{equation*}
%
This form of soft update addresses uncertainty through scheduling an increasing $\beta$ with more training as a result of decreased model uncertainty. The mellowmax operator smoothly degrades to max when model uncertainty is lowered during training. This allows convergence of our method to the same optimal fix point $Q^*$ as Q-Learning.

Combining the above analysis, following \cite{Fox2016Taming, pmlr-v70-asadi17a} we can then show, 
\begin{lemma} \label{bellman_contraction_lemma}
    The Bellman operator $\C{B}_w$ is a contraction between $Q^{(i)}$ and $Q^{(j)}$ for all $i, j$ in $L^\infty$: 
    $$\| \C{B}_w[Q^{(i)}] - \C{B}_w[Q^{(j)}] \|_\infty \le \gamma \| Q^{(i)} - Q^{(j)} \|_\infty.$$
\end{lemma} \vspace{-\baselineskip}
\begin{proof} 
  See Appendix~\ref{bellman_contraction}. 
\end{proof}
In practice, we update the ensemble members $Q^{(k)}$ using a stochastic approximation to the Bellman operator 
resulting from our observed reward $r$ and new state $s'$.
We therefore adapt the proof of convergence for Q-learning from \cite{jaakkola1994convergence} to show that,
%
%
%
\begin{theorem} \label{UQL_convergence}
    If $S$ and $A$ are finite sets, and the learning rate $\alpha_t(s,a)$ obeys $\sum_t \alpha_t(s,a) = \infty$ and $\sum_t \alpha_t^2(s,a) < \infty$,
    then the Unbiased Soft Q-Learning ensemble converges to $Q^{(i)} = Q^{(j)} = Q^*$ for all $i,j$ with probability one 
    via the stochastic update rule in \Eqref{UQL_update}. 
\end{theorem} \vspace{-\baselineskip}
\begin{proof}
See Appendix~\ref{UQL_convergence_proof}.
\end{proof}
The full algorithm is given in Algorithm~\ref{pseudo_code}. 

\begin{algorithm}[H]
         Initialize replay memory $D$ to capacity $N$ \; \\
         Initialize $K$ parameterized state-action value function $\{Q^{(k)}\}$; save $\bar{Q}^{(k)} = Q^{(k)}$ for all $k$ \\
         Initialize $\beta$ search range $B = [a, b]$, correction constant $\kappa$ \\
         \For{$t=1, \ldots, T$}{
          Select action $a_t$ according to arbitrary policy $\pi$ \; \newline
          Execute action $a_t$ and observe reward $r_t$ and next state observation $s_{t+1}$ \; \\
          Store transition $(s_t, a_t, r_t, s_{t+1})$ in $D$ \; \\
          $s_{t} \leftarrow s_{t+1}$ \; \\
          \For{$k=1, \ldots, K$} 
              {Sample random minibatch of transitions $(s_j, a_j, r_j, s_j')$ from $D$ \; \\
              $f_{Q,s}(\beta) = \hat{\E}_{Q}\big[V_{\beta}(s)\big]- \max_a \hat{\E}_{Q}\big[Q(s, a)\big]$ \; \\
              $\beta_{s_j'} \leftarrow \text{BinarySearch}(f_{s_j';\bar{Q}}, B)$ \; \\
              $y_{j}^{(k)}= \begin{cases}
              r_j, & \text{if $s_{j}'$ is terminal state}. \\
              r_j + \gamma V^{(k)}_{\kappa\beta_{s_{j}'}}(s_{j}'), & \text{otherwise}.
              \end{cases} $\\
            Update $Q^{(k)}$ with TD error $y_{j}^{(k)}-Q^{(k)}(s_j, a_j)$ \; }
          Every once in a while: $\bar{Q}^{(k)} \leftarrow Q^{(k)}$ for all $k$ \;
         } 
         
         \caption{Unbiased Soft Q-Learning with ensemble of approximators} \label{pseudo_code}
\end{algorithm}

\subsection{Visualizing the Estimation Bias in Early Stage Training}
\begin{figure}[t]
\begin{centering}
\subcaptionbox{\label{gridworld:a}}{\includegraphics[width=0.305\textwidth]{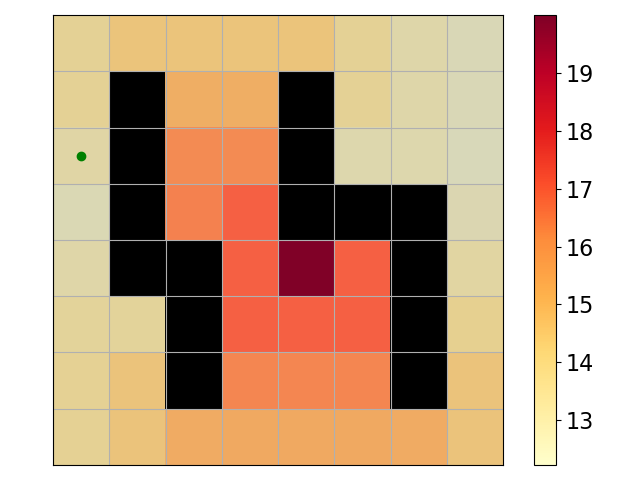}}
\subcaptionbox{\label{gridworld:b}}{\includegraphics[width=0.33\textwidth]{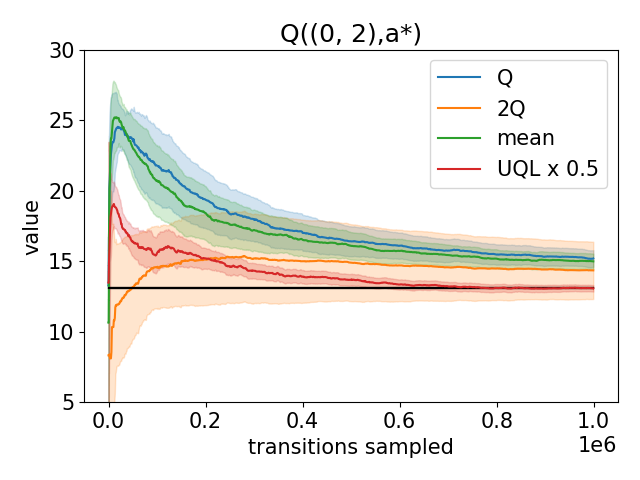}}
\subcaptionbox{\label{gridworld:c}}{\includegraphics[width=0.33\textwidth]{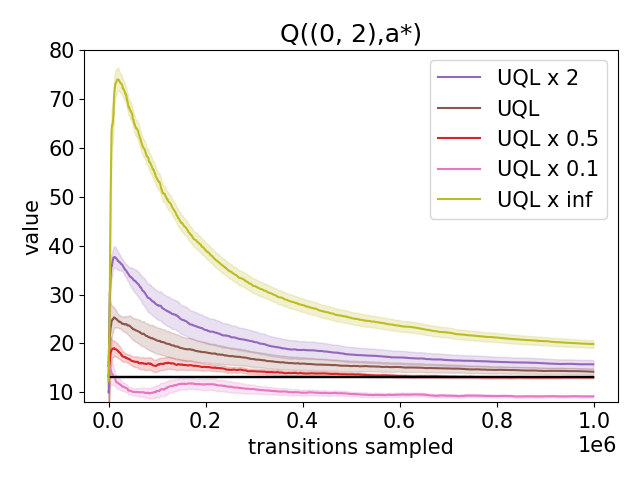}}
\end{centering}
\caption{(a) Optimal state value in the grid world domain. Walls (black) are impassable. The absorbing goal state has the highest value. (b) Value estimation (colored) of Q-Learning (Q), Double Q-Learning (2Q), using ensemble averaged Q for update (mean), UQL with $\kappa=0.5$, and the ground truth value (black) of the state (0, 2) is marked by the left green dot, averaged over 20 ensemble estimates of 10 runs. Standard deviation (shaded) is over the estimation of different runs. (c) same as (b) but with different $\kappa \in \{1, 2, 0.5, 0.1, \infty\}$.}
\label{gridworld}
\end{figure}

To analyze the status of this learning algorithm at each specific state, we tested our method in a Gridworld domain (Figure \ref{gridworld:a}). The black tiles are walls; colored tiles are valid states. At each state, the agent may move in any of 8 different directions. Then, environment dynamics may cause the agent to slip into to a neighboring state of the desired one with some probability.  We maintain an ensemble of Q tables for our method. To eliminate the influence of an exploration policy, we perform stochastic updates on the Q ensemble with uniformly sampled $s$ and $a$ and then sample $r$ and $s'$ according to the MDP dynamics. 

This toy example highlights the difference between Q-Learning and Unbiased Soft Q-Learning in terms of reducing bias propagation. In Figure \ref{gridworld:b}, we see that Q-Learning overestimates the value. After estimating the inverse temperature $\beta$, and updating each ensemble member with correction $\kappa=0.5$, 
the estimation bias is significantly decreased in early stage training, leading to faster convergence. 
The positive bias effect increases with the value of $\kappa$ used in the update (see Figure \ref{gridworld:c}).

\subsection{Policy Quality in Early Stage Training}
\begin{figure}[h]
\begin{center}
\subcaptionbox{Optimal Policy\label{policy:a}}{\includegraphics[width=0.31\textwidth]{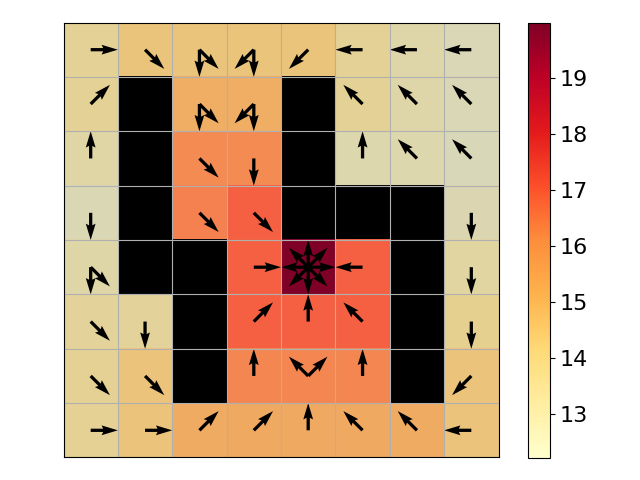}}
\subcaptionbox{Q-Learning \label{policy:b}}{\includegraphics[width=0.31\textwidth]{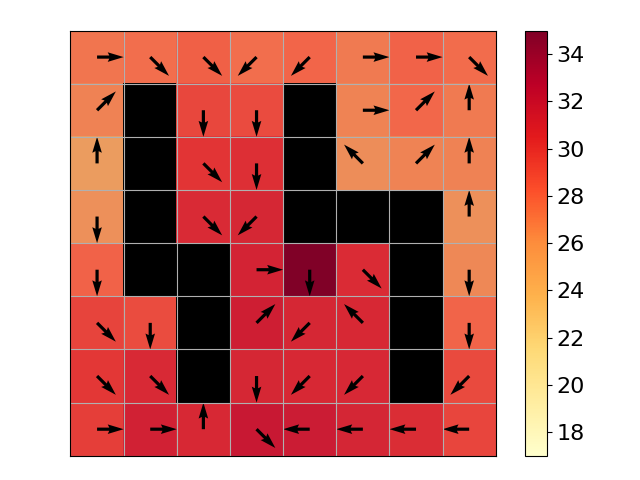}}
\subcaptionbox{UQL$\times$0.5 \label{policy:c}}{\includegraphics[width=0.31\textwidth]{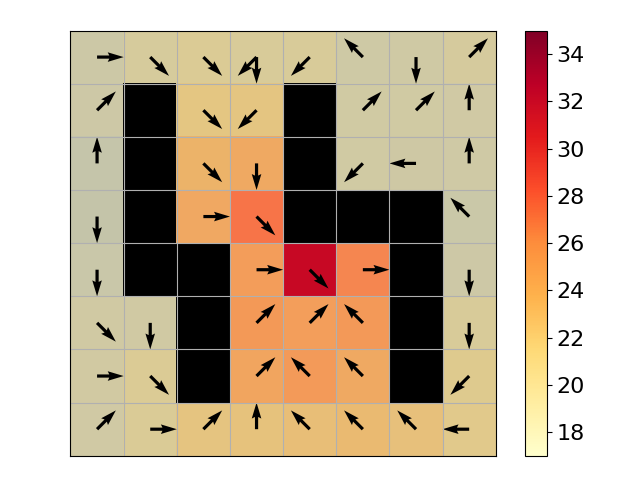}}
%
%
\end{center}
\caption{(a) True value $V^*(s)$ and optimal policy $\pi^*(s)$ of each Gridworld state.
(b)--(c) Estimated value function $V(s)=\max_a \hat{\E}_Q[Q^{(k)}(s,a)]$ and greedy policy $\pi(s) = \argmax_a \hat{\E}_Q[Q^{(k)}(s,a)]$
for (b) standard Q-learning, and (c) UQL with $\kappa=0.5$, after 10000 stochastic updates.
}
\end{figure}
Reducing the bias in our estimates of $Q$ can also significantly improve our estimated policy during training.
In particular, since the max operator of Q-Learning overestimates the value in early stages, the positive bias may swamp the true relative
values of subsequent states, making the policy converge more slowly to its optimum. 
For example, in the Gridworld domain, we can see that, in addition to the improved quality of the value estimates of 
each state, the estimated policy of most states also converges faster for UQL (Figure \ref{policy:c}) than using 
standard Q-Learning (Figure \ref{policy:b}).


\section{Learning in High-Dimensional Continuous State Space MDPs}
In MDPs with extremely many or even continuous state values, one can approximate the state-action value function $Q$ using
a parameterized model such as a deep neural network (DNN).  Our approach naturally extends to this environment by using ensembles of neural approximations.

\subsection{Unbiased Soft Q-Learning with Ensemble of Function Approximators}
We use an ensemble of DNNs to learn to approximate Q values of each high-dimensional state-action pair (Algorithm \ref{pseudo_code}), which characterizes the uncertainty of value predictions. As is typical, our loss function is the mean squared Bellman error of a batch of transitions sampled from the replay buffer. Previous methods \citep{lan2020maxmin, DBLP:journals/corr/abs-2007-04938} have shown the effectiveness of ensemble learning in combating the bias in value estimation. \citet{lan2020maxmin} takes the minimum estimate over the ensemble in order to learn the value more pessimistically. Ensemble methods come with a natural notion of model uncertainty, often reflected through prediction entropy or variance. An example is \citet{DBLP:journals/corr/abs-2007-04938}, which improves sample efficiency by updating the value estimator with a Bellman operator weighted by a quantified estimation of model uncertainty, which is a value related to the output variance of Q networks. In this paper, we use an ensemble to approximate the model output distribution. We update the ensemble iteratively according to the Unbiased Soft Update \Eqref{UQL_update}, and compare the results with widely used update rules. 
%
%
%

%
\subsection{Additional Improvements}
    To further improve the performance of UQL, we included several helpful adaptations that fit easily in our ensemble Soft Q-Learning framework. Upper Confidence Exploration \citep{audibert2009exploration, auer2002finite}, Prioritized Experience Replay \citep{schaul2015prioritized}, and Dueling Networks \citep{wang2016dueling}. There are also methods that work well, and are used in Rainbow DQN \citep{hessel2018rainbow}, that are not included in our method, such as Double DQN \citep{DBLP:journals/corr/HasseltGS15}, Distributional Q \citep{bellemare2017distributional}, Multi-step value estimation \citep{sutton1988learning, sutton2018reinforcement}, and Noisy networks \citep{DBLP:journals/corr/FortunatoAPMOGM17}. It remains an open question if these can be incorporated to improve UQL.

\section{Experiments}
In order to understand the effect of unbiased estimation in theoretical and practical settings, we evaluate our unbiased update in Atari environments with neural network Q function representations. We design our experiments to answer the following questions:
\begin{itemize}\setlength\itemsep{0em}
    \item Can unbiased soft updates improve performance?
    \item How well does unbiased soft update control value estimation bias?
    \item Does temperature reflect model uncertainty?
    \item How do different correction constants affect performance and bias?
\end{itemize}

    \begin{wrapfigure}{r}{0.5\textwidth}
        \centering
        \vspace{-20pt}
        \hspace*{-1.4cm}\includegraphics[width=0.73\textwidth]{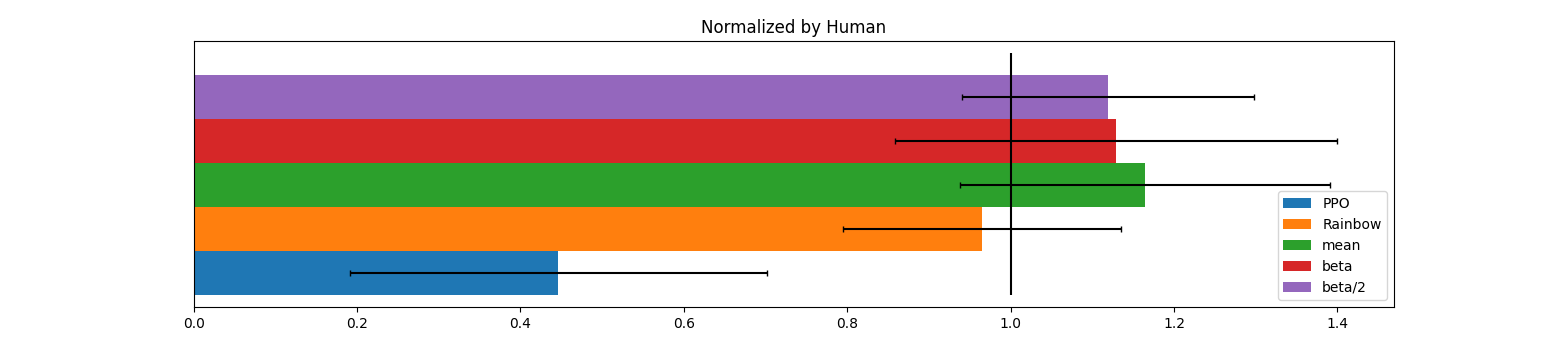}
        \caption{Performance score achieved after 500k interactions, averaged over the 25 Atari environments in Appendix \ref{main_results}. Performance score of each environment is computed using: $(\frac{\text{Score}_{\text{Algorithm}} - \text{Score}_{\text{Random}}}{\text{Score}_{\text{Human}} - \text{Score}_{\text{Random}}})$. UQL with $\kappa\in\{1, 0.5\}$ is shown in red and purple.}
        \label{normalized_measure}
        \vspace{-25pt}
    \end{wrapfigure}
    
    \subsection{Overall Performance}
    Figure~\ref{normalized_measure} shows the average performance of UQL across 25 Atari environments, normalized by random (0) and human performance scores (1).
    PPO and Rainbow DQN values are as reported in \cite{DBLP:journals/corr/abs-1903-00374}.
    Detailed performance values are given in in Appendix \ref{main_results}.
    
    \subsection{Comparison to Rainbow DQN}
    UQL executes 1 update for each ensemble member for every 2 transitions sampled from the environment. The temperature used in this experiment is not accounting for finite ensemble size. A modified version of Rainbow, which updates with the exact same frequency (5 updates for every 2 transitions sampled) as UQL. This shows that the performance increase from more frequent gradient update is limited. Results in Appendix \ref{ablation}.  
    
    \subsection{The effect of unbiased soft update}
    
    \begin{figure}[t] 
        \centering
        \begin{center}
        \includegraphics[width=0.85\textwidth]{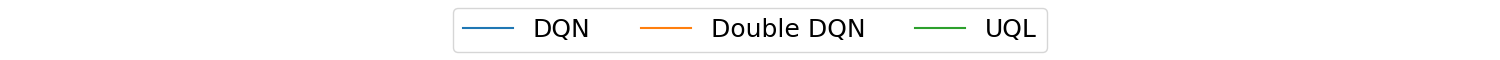} \\
        \includegraphics[width=0.24\textwidth]{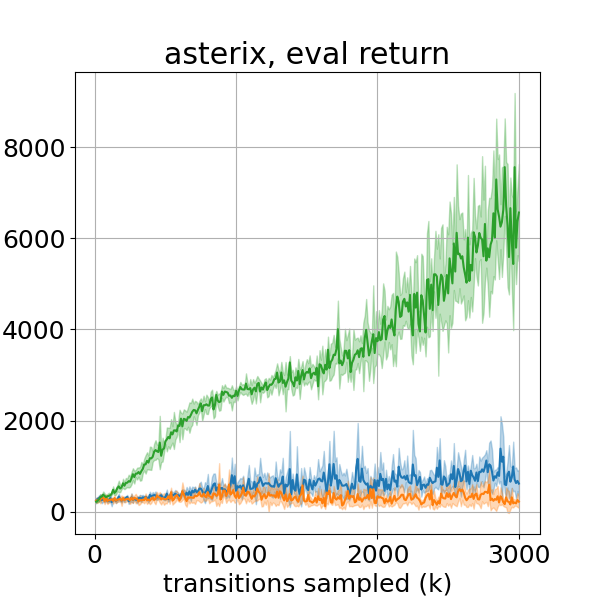}
        \includegraphics[width=0.24\textwidth]{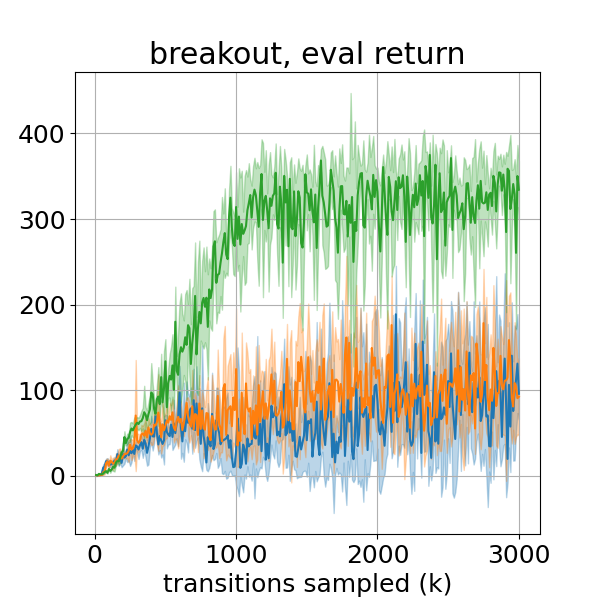}
        \includegraphics[width=0.24\textwidth]{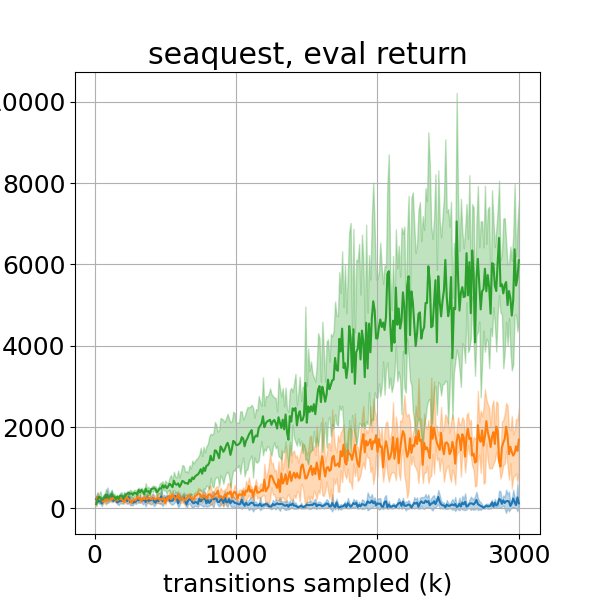}
        \includegraphics[width=0.24\textwidth]{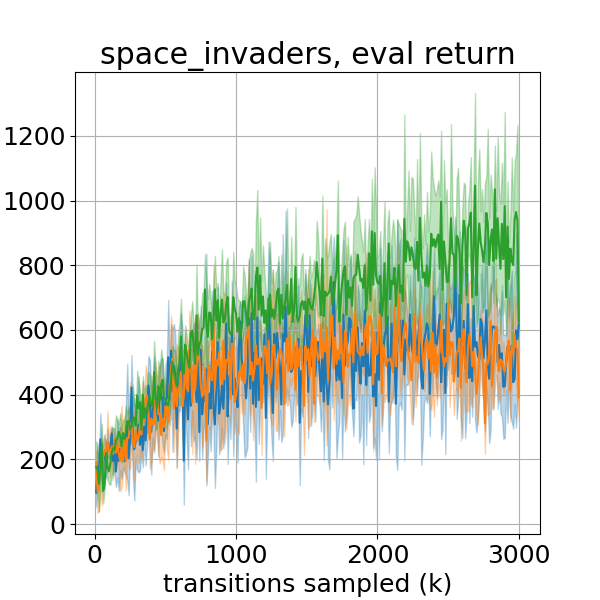} \\
        \includegraphics[width=0.24\textwidth]{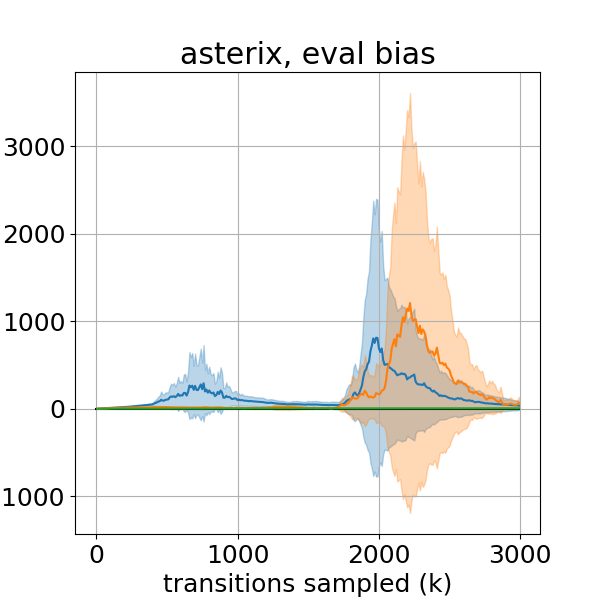}
        \includegraphics[width=0.24\textwidth]{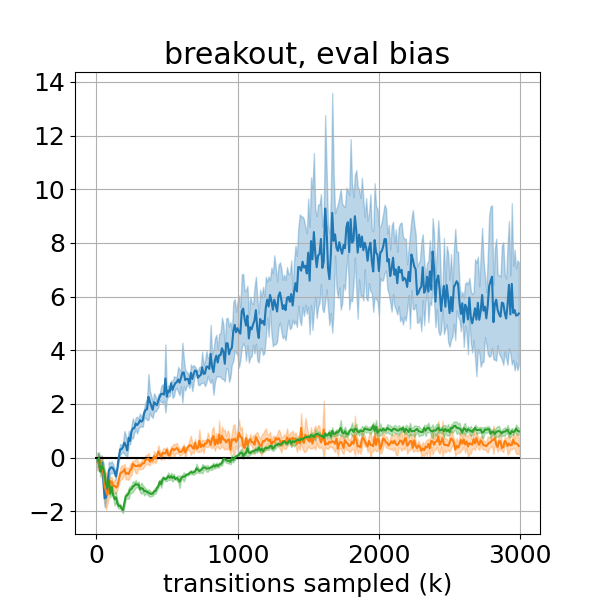}
        \includegraphics[width=0.24\textwidth]{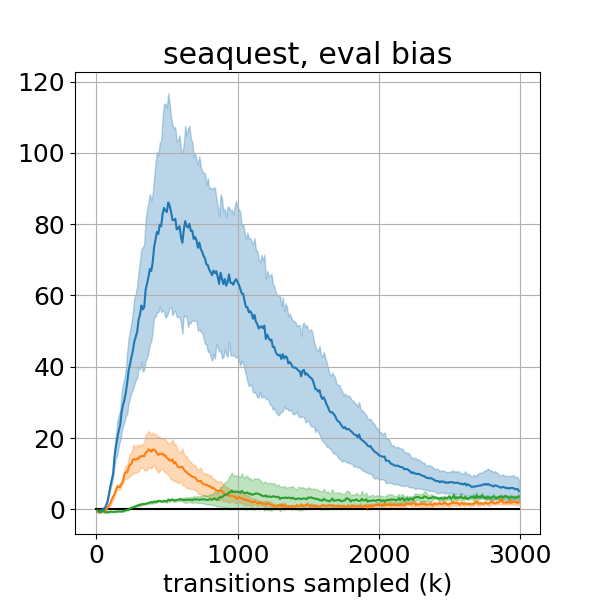}
        \includegraphics[width=0.24\textwidth]{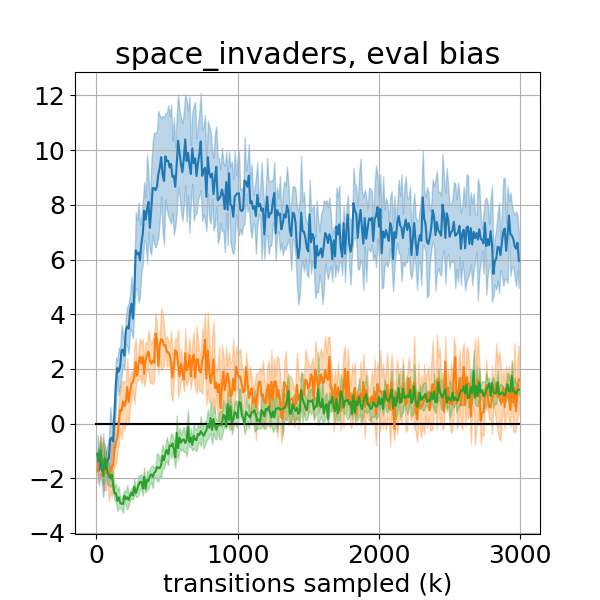}
        \end{center}

        \caption{First row: Average score over 5 runs. Second row: Initial state discounted value estimation bias averaged over 5 runs. Error bar is the standard deviation of the reported value over 5 runs. Prioritized replay, Dueling Networks are not used in these experiments.}\label{sep_replay_uql}
    \end{figure}
    
    To visualize the accuracy of the value prediction that unbiased soft updates gives and the quality of the trained policy compare to other update methods for learning Q networks, we experimented in 4 Atari domains (shown in Figure \ref{sep_replay_uql}). We use $\hat{\E}_{\xi\sim p_{\pi_\theta}(\xi)}[V_\theta(s_0) - R(\xi)]$ as a proxy to the overall estimation bias of the Q networks. We can see that in this experiment, lower estimation is linked to better performance.
    
    To isolate the effect of the exploration strategy, these experiments use an experimental setting that shares the replay buffer between algorithms, in order to eliminate any influence on the exploration quality induced by learned Q values. In each training iteration, each algorithms samples 1 transition using epsilon-greedy exploration policy from a copy (assigned to that algorithm) of the Atari game, and stores to the shared replay buffer. Then, each algorithm samples a batch uniformly from the shared replay buffer and takes a gradient step on its Q network using target estimates generated by that algorithm. UQL is still able to outperform in terms of performance and bias, results shown in Appendix \ref{ablation}.
    
    \subsection{Unbiased Temperature and Model Uncertainty}
    
    \begin{figure}[h] 
        \centering
        
        
        \subcaptionbox{$\log w = -5.2$\label{temp_uql:a}}{\includegraphics[width=0.24\textwidth]{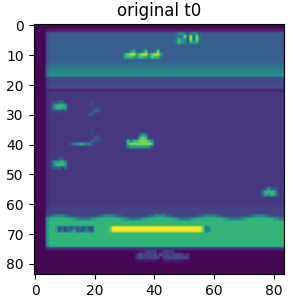}}
        \subcaptionbox{$\log w = -5.1$\label{temp_uql:b}}{\includegraphics[width=0.24\textwidth]{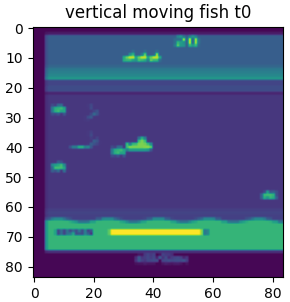}}
        \subcaptionbox{$\log w = -4.1$\label{temp_uql:c}}{\includegraphics[width=0.24\textwidth]{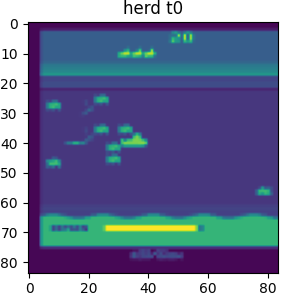}}
        \subcaptionbox{$\log w = -1.7$\label{temp_uql:d}}{\includegraphics[width=0.24\textwidth]{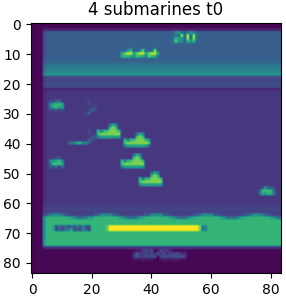}}
        \caption{Awareness of novel state through unbiased temperature scheduling. $w = \frac{1}{\beta_{Q_\theta}(s)}$.}
        \label{temp_uql}
    \end{figure}
    
    To visualize the state unbiased temperature, we perturbed a normal state representation framestack of Seaquest (Figure \ref{temp_uql:a}) into a state that the learned agent has never seen (Figure \ref{temp_uql:b}, \ref{temp_uql:c}, \ref{temp_uql:d}) with increasing insanity. We compute the temperature of each state using a model trained with UQL for 5.1M steps in these 4 states. The action distribution selected to evaluate the target estimate will be closer to an $\argmax$ over action in Figure \ref{temp_uql:a}  with which the agent is familiar with, and closer to uniform in Figure \ref{temp_uql:d} with which the agent is unfamiliar.
    
    \subsection{Using a Different Correction Constant}
    \begin{figure}[h]
        \centering
        \includegraphics[width=0.85\textwidth] {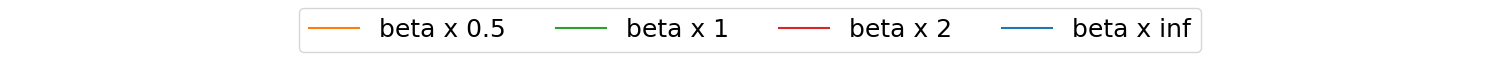} \\
        \includegraphics[width=0.19\textwidth]{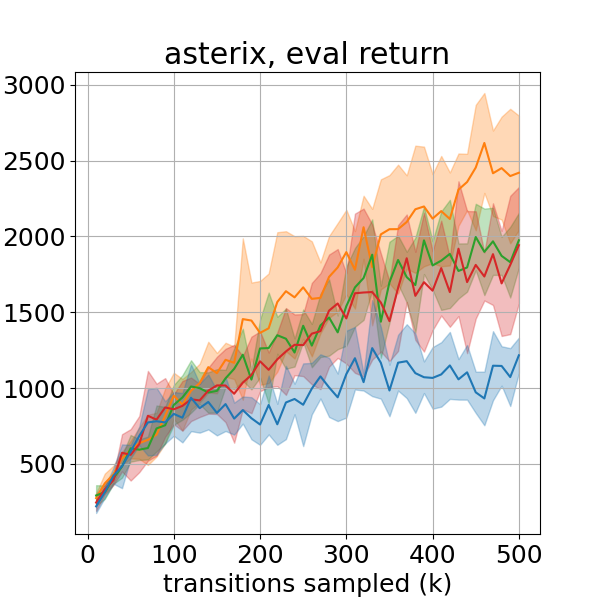}
        \includegraphics[width=0.19\textwidth]{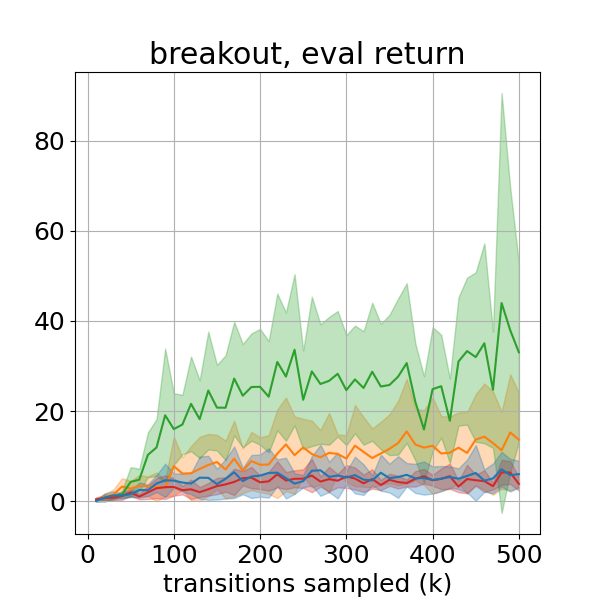}
        \includegraphics[width=0.19\textwidth]{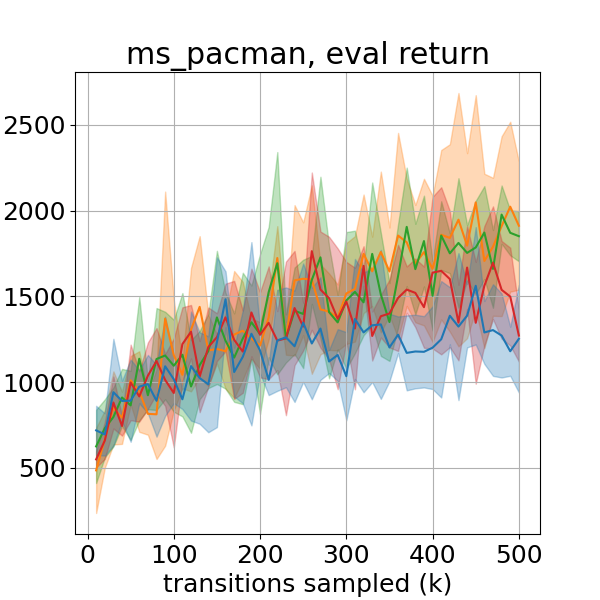}
        \includegraphics[width=0.19\textwidth]{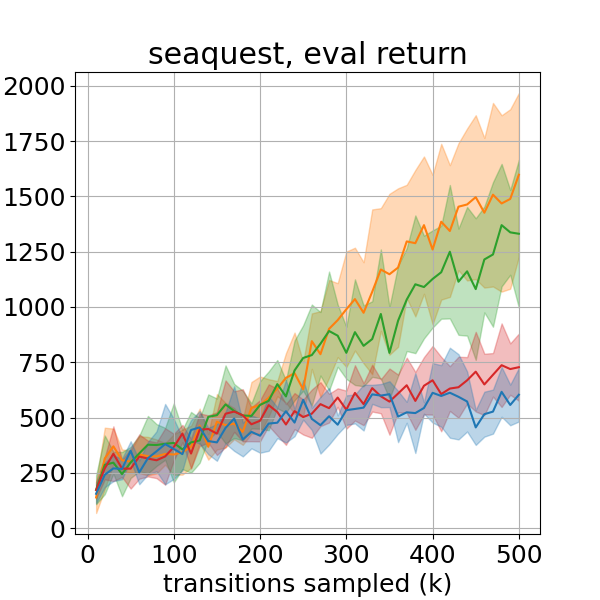}
        \includegraphics[width=0.19\textwidth]{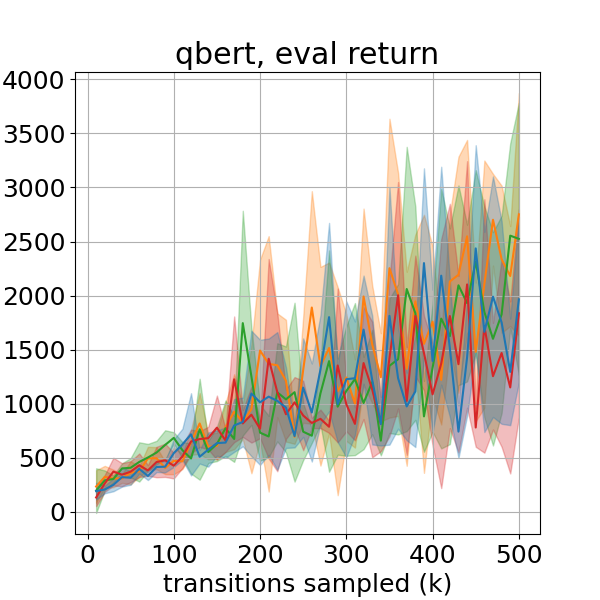} \\
        \includegraphics[width=0.19\textwidth]{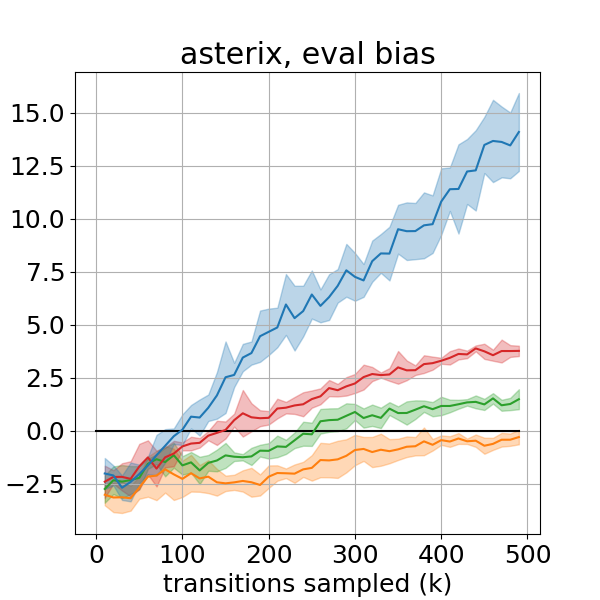}
        \includegraphics[width=0.19\textwidth]{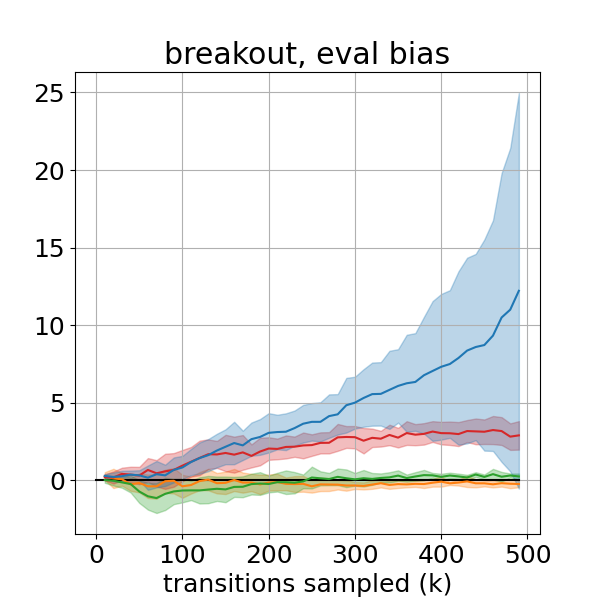}
        \includegraphics[width=0.19\textwidth]{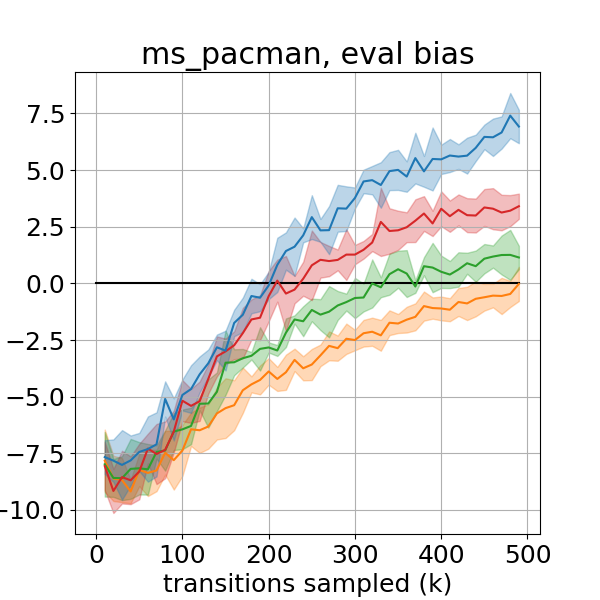}
        \includegraphics[width=0.19\textwidth]{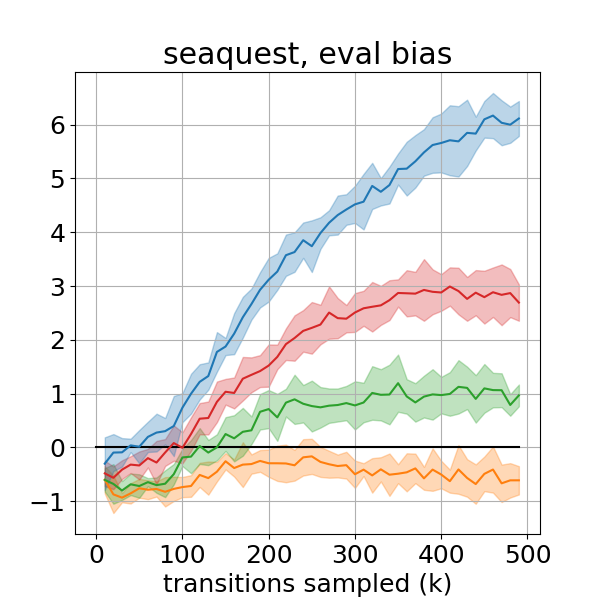}
        \includegraphics[width=0.19\textwidth]{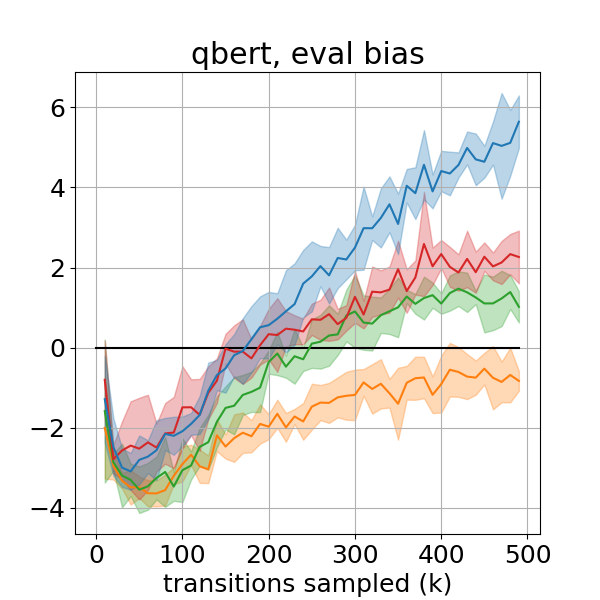}
        \caption{First row: Average score of 1 life over 5 runs, experience is not shared between algorithms. Second row: Initial state discounted value estimation bias averaged over 5 runs. Error bar is the standard deviation of the reported value over 5 runs.}
        \label{fig:inaccurate_temp_return_bias}
    \end{figure}
    
    In Figure \ref{fig:inaccurate_temp_return_bias}, we run the algorithm with $w = ({\kappa\beta_{Q_\theta}(s)})^{-1}$ as the temperature for update with $\kappa \in \{\infty, 2, 1, 0.5\}$ on 5 ensemble members. Q-Learning is equivalent to using $\kappa=\infty$. From this experiment, we can see that bias is monotonic in $\beta$. A correction constant $\kappa\in(0,1)$ may be tuned for performance to counter overestimation of Q from a finite collection of approximators. We can see that lower bias is linked to better performance, but the strength of this correlation can be affected by other factors of training like exploration.

    
    \subsection{Using a Different Next-State Reduction Operator}
    In Figure \ref{fig:uql_diff_operator}, we use an alternative target estimates. Softmax of the next state $V_\beta(s)=\E_{a\sim\pi_\beta(\cdot|s)}[Q(s,a)]$, which evaluates the next state value as the expected Q value following Boltzman distribution over actions ($\pi_\beta(a|s)\defeq\frac{\pi_0(a|s)\exp(\beta Q(s,a))}{\sum_{\bar{a}}\pi_0(\bar{a}|s)\exp(\beta Q(s,\bar{a}))}$). We use these estimators to substitute the maximum-entropy target estimate and search the parameter $\beta$ updates. The baseline method ``mean" updates the ensemble member $Q^{(i)}$ with TD error $r + \gamma \max_a \hat{\E}_Q[Q(s', a')] - Q^{(i)}(s, a)$.



\section{Discussion and Future Work}
UQL is based on the temperature scheduling of EQL \citep{Fox2019Toward}, but gives a tractable numerical solution to find the temperature of the maximum-entropy target estimate used in SQL, which is more theoretically motivated than EQL. In this work, we correct the bias from estimating the value using $\max_a \hat{\E}_Q[Q(s,a)]$ with $\kappa$, which can be environment and exploration dependent. Therefore, future work on a better approximation of the unbiased target will be potentially helpful to derive a better algorithm. It is also clear that less estimation bias does not necessarily lead to better performance, which depends not only the value estimation, but also on the exploration strategy. It will be also beneficial to have more theoretical results on the properties of an exploration strategy during its interaction with a noisy estimate of Q. It would also be interesting to conduct experiments that show the reason why just using ensemble estimate mean in neural network Q representation works well already. It is obvious that the success of this method comes partially from this mean target estimate. 



\bibliography{main}

\begin{thebibliography}{31}
\providecommand{\natexlab}[1]{#1}
\providecommand{\url}[1]{\texttt{#1}}
\expandafter\ifx\csname urlstyle\endcsname\relax
  \providecommand{\doi}[1]{doi: #1}\else
  \providecommand{\doi}{doi: \begingroup \urlstyle{rm}\Url}\fi

\bibitem[Anonymous1(2021)]{countbased2021}
Anonymous1.
\newblock Count-based temperature scheduling for maximum entropy reinforcement
  learning.
\newblock \emph{in submission}, 2021.

\bibitem[Anonymous2(2021)]{TESSAC2021}
Anonymous2.
\newblock Target entropy annealing for discrete soft actor-critic.
\newblock \emph{in submission}, 2021.

\bibitem[Asadi \& Littman(2017)Asadi and Littman]{pmlr-v70-asadi17a}
Kavosh Asadi and Michael~L. Littman.
\newblock An alternative softmax operator for reinforcement learning.
\newblock In Doina Precup and Yee~Whye Teh (eds.), \emph{Proceedings of the
  34th International Conference on Machine Learning}, volume~70 of
  \emph{Proceedings of Machine Learning Research}, pp.\  243--252. PMLR, 06--11
  Aug 2017.
\newblock URL \url{https://proceedings.mlr.press/v70/asadi17a.html}.

\bibitem[Audibert et~al.(2009)Audibert, Munos, and
  Szepesv{\'a}ri]{audibert2009exploration}
Jean-Yves Audibert, R{\'e}mi Munos, and Csaba Szepesv{\'a}ri.
\newblock Exploration--exploitation tradeoff using variance estimates in
  multi-armed bandits.
\newblock \emph{Theoretical Computer Science}, 410\penalty0 (19):\penalty0
  1876--1902, 2009.

\bibitem[Auer et~al.(2002)Auer, Cesa-Bianchi, and Fischer]{auer2002finite}
Peter Auer, Nicolo Cesa-Bianchi, and Paul Fischer.
\newblock Finite-time analysis of the multiarmed bandit problem.
\newblock \emph{Machine learning}, 47\penalty0 (2):\penalty0 235--256, 2002.

\bibitem[Bellemare et~al.(2017)Bellemare, Dabney, and
  Munos]{bellemare2017distributional}
Marc~G Bellemare, Will Dabney, and R{\'e}mi Munos.
\newblock A distributional perspective on reinforcement learning.
\newblock In \emph{International Conference on Machine Learning}, pp.\
  449--458. PMLR, 2017.

\bibitem[Fortunato et~al.(2017)Fortunato, Azar, Piot, Menick, Osband, Graves,
  Mnih, Munos, Hassabis, Pietquin, Blundell, and
  Legg]{DBLP:journals/corr/FortunatoAPMOGM17}
Meire Fortunato, Mohammad~Gheshlaghi Azar, Bilal Piot, Jacob Menick, Ian
  Osband, Alex Graves, Vlad Mnih, R{\'{e}}mi Munos, Demis Hassabis, Olivier
  Pietquin, Charles Blundell, and Shane Legg.
\newblock Noisy networks for exploration.
\newblock \emph{CoRR}, abs/1706.10295, 2017.
\newblock URL \url{http://arxiv.org/abs/1706.10295}.

\bibitem[Fox(2019)]{Fox2019Toward}
Roy Fox.
\newblock Toward provably unbiased temporal-difference value estimation.
\newblock In \emph{Optimization Foundations for Reinforcement Learning workshop
  (OPTRL @ NeurIPS)}, 2019.

\bibitem[Fox et~al.(2016)Fox, Pakman, and Tishby]{Fox2016Taming}
Roy Fox, Ari Pakman, and Naftali Tishby.
\newblock Taming the noise in reinforcement learning via soft updates.
\newblock In \emph{32nd Conference on Uncertainty in Artificial Intelligence
  (UAI)}, 2016.

\bibitem[Haarnoja et~al.(2017)Haarnoja, Tang, Abbeel, and
  Levine]{DBLP:journals/corr/HaarnojaTAL17}
Tuomas Haarnoja, Haoran Tang, Pieter Abbeel, and Sergey Levine.
\newblock Reinforcement learning with deep energy-based policies.
\newblock \emph{CoRR}, abs/1702.08165, 2017.
\newblock URL \url{http://arxiv.org/abs/1702.08165}.

\bibitem[Haarnoja et~al.(2018)Haarnoja, Zhou, Hartikainen, Tucker, Ha, Tan,
  Kumar, Zhu, Gupta, Abbeel, et~al.]{haarnoja2018soft_app}
Tuomas Haarnoja, Aurick Zhou, Kristian Hartikainen, George Tucker, Sehoon Ha,
  Jie Tan, Vikash Kumar, Henry Zhu, Abhishek Gupta, Pieter Abbeel, et~al.
\newblock Soft actor-critic algorithms and applications.
\newblock \emph{arXiv preprint arXiv:1812.05905}, 2018.

\bibitem[Hessel et~al.(2018)Hessel, Modayil, Van~Hasselt, Schaul, Ostrovski,
  Dabney, Horgan, Piot, Azar, and Silver]{hessel2018rainbow}
Matteo Hessel, Joseph Modayil, Hado Van~Hasselt, Tom Schaul, Georg Ostrovski,
  Will Dabney, Dan Horgan, Bilal Piot, Mohammad Azar, and David Silver.
\newblock Rainbow: Combining improvements in deep reinforcement learning.
\newblock In \emph{Thirty-second AAAI conference on artificial intelligence},
  2018.

\bibitem[Jaakkola et~al.(1994)Jaakkola, Jordan, and
  Singh]{jaakkola1994convergence}
Tommi Jaakkola, Michael~I Jordan, and Satinder~P Singh.
\newblock On the convergence of stochastic iterative dynamic programming
  algorithms.
\newblock \emph{Neural computation}, 6\penalty0 (6):\penalty0 1185--1201, 1994.

\bibitem[Kaiser et~al.(2019)Kaiser, Babaeizadeh, Milos, Osinski, Campbell,
  Czechowski, Erhan, Finn, Kozakowski, Levine, Sepassi, Tucker, and
  Michalewski]{DBLP:journals/corr/abs-1903-00374}
Lukasz Kaiser, Mohammad Babaeizadeh, Piotr Milos, Blazej Osinski, Roy~H.
  Campbell, Konrad Czechowski, Dumitru Erhan, Chelsea Finn, Piotr Kozakowski,
  Sergey Levine, Ryan Sepassi, George Tucker, and Henryk Michalewski.
\newblock Model-based reinforcement learning for atari.
\newblock \emph{CoRR}, abs/1903.00374, 2019.
\newblock URL \url{http://arxiv.org/abs/1903.00374}.

\bibitem[Kumar et~al.(2019)Kumar, Fu, Soh, Tucker, and
  Levine]{NEURIPS2019_c2073ffa}
Aviral Kumar, Justin Fu, Matthew Soh, George Tucker, and Sergey Levine.
\newblock Stabilizing off-policy q-learning via bootstrapping error reduction.
\newblock In H.~Wallach, H.~Larochelle, A.~Beygelzimer, F.~d\textquotesingle
  Alch\'{e}-Buc, E.~Fox, and R.~Garnett (eds.), \emph{Advances in Neural
  Information Processing Systems}, volume~32. Curran Associates, Inc., 2019.
\newblock URL
  \url{https://proceedings.neurips.cc/paper/2019/file/c2073ffa77b5357a498057413bb09d3a-Paper.pdf}.

\bibitem[Kumar et~al.(2020)Kumar, Gupta, and Levine]{NEURIPS2020_d7f426cc}
Aviral Kumar, Abhishek Gupta, and Sergey Levine.
\newblock Discor: Corrective feedback in reinforcement learning via
  distribution correction.
\newblock In H.~Larochelle, M.~Ranzato, R.~Hadsell, M.~F. Balcan, and H.~Lin
  (eds.), \emph{Advances in Neural Information Processing Systems}, volume~33,
  pp.\  18560--18572. Curran Associates, Inc., 2020.
\newblock URL
  \url{https://proceedings.neurips.cc/paper/2020/file/d7f426ccbc6db7e235c57958c21c5dfa-Paper.pdf}.

\bibitem[Lan et~al.(2020)Lan, Pan, Fyshe, and White]{lan2020maxmin}
Qingfeng Lan, Yangchen Pan, Alona Fyshe, and Martha White.
\newblock Maxmin q-learning: Controlling the estimation bias of q-learning,
  2020.

\bibitem[Lee et~al.(2020)Lee, Laskin, Srinivas, and
  Abbeel]{DBLP:journals/corr/abs-2007-04938}
Kimin Lee, Michael Laskin, Aravind Srinivas, and Pieter Abbeel.
\newblock {SUNRISE:} {A} simple unified framework for ensemble learning in deep
  reinforcement learning.
\newblock \emph{CoRR}, abs/2007.04938, 2020.
\newblock URL \url{https://arxiv.org/abs/2007.04938}.

\bibitem[Liu \& Ihler(2011)Liu and Ihler]{liu2011bounding}
Qiang Liu and Alexander~T Ihler.
\newblock Bounding the partition function using holder's inequality.
\newblock In \emph{ICML}, 2011.

\bibitem[Mnih et~al.(2013)Mnih, Kavukcuoglu, Silver, Graves, Antonoglou,
  Wierstra, and Riedmiller]{DBLP:journals/corr/MnihKSGAWR13}
Volodymyr Mnih, Koray Kavukcuoglu, David Silver, Alex Graves, Ioannis
  Antonoglou, Daan Wierstra, and Martin~A. Riedmiller.
\newblock Playing atari with deep reinforcement learning.
\newblock \emph{CoRR}, abs/1312.5602, 2013.
\newblock URL \url{http://arxiv.org/abs/1312.5602}.

\bibitem[Mnih et~al.(2015)Mnih, Kavukcuoglu, Silver, Rusu, Veness, Bellemare,
  Graves, Riedmiller, Fidjeland, Ostrovski, et~al.]{mnih2015human}
Volodymyr Mnih, Koray Kavukcuoglu, David Silver, Andrei~A Rusu, Joel Veness,
  Marc~G Bellemare, Alex Graves, Martin Riedmiller, Andreas~K Fidjeland, Georg
  Ostrovski, et~al.
\newblock Human-level control through deep reinforcement learning.
\newblock \emph{nature}, 518\penalty0 (7540):\penalty0 529--533, 2015.

\bibitem[Rubin et~al.(2012)Rubin, Shamir, and Tishby]{rubin2012trading}
Jonathan Rubin, Ohad Shamir, and Naftali Tishby.
\newblock Trading value and information in mdps.
\newblock In \emph{Decision Making with Imperfect Decision Makers}, pp.\
  57--74. Springer, 2012.

\bibitem[Schaul et~al.(2015)Schaul, Quan, Antonoglou, and
  Silver]{schaul2015prioritized}
Tom Schaul, John Quan, Ioannis Antonoglou, and David Silver.
\newblock Prioritized experience replay.
\newblock \emph{arXiv preprint arXiv:1511.05952}, 2015.

\bibitem[Schulman et~al.(2017)Schulman, Wolski, Dhariwal, Radford, and
  Klimov]{schulman2017proximal}
John Schulman, Filip Wolski, Prafulla Dhariwal, Alec Radford, and Oleg Klimov.
\newblock Proximal policy optimization algorithms.
\newblock \emph{arXiv preprint arXiv:1707.06347}, 2017.

\bibitem[Sutton(1988)]{sutton1988learning}
Richard~S Sutton.
\newblock Learning to predict by the methods of temporal differences.
\newblock \emph{Machine learning}, 3\penalty0 (1):\penalty0 9--44, 1988.

\bibitem[Sutton \& Barto(2018)Sutton and Barto]{sutton2018reinforcement}
Richard~S Sutton and Andrew~G Barto.
\newblock \emph{Reinforcement learning: An introduction}.
\newblock MIT press, 2018.

\bibitem[Thrun \& Schwartz(1993)Thrun and Schwartz]{Thrun-1993-15908}
Sebastian Thrun and A.~Schwartz.
\newblock Issues in using function approximation for reinforcement learning.
\newblock In M.~Mozer, P.~Smolensky, D.~Touretzky, J.~Elman, and A.~Weigend
  (eds.), \emph{Proceedings of the 1993 Connectionist Models Summer School}.
  Erlbaum Associates, June 1993.

\bibitem[van Hasselt et~al.(2015)van Hasselt, Guez, and
  Silver]{DBLP:journals/corr/HasseltGS15}
Hado van Hasselt, Arthur Guez, and David Silver.
\newblock Deep reinforcement learning with double q-learning.
\newblock \emph{CoRR}, abs/1509.06461, 2015.
\newblock URL \url{http://arxiv.org/abs/1509.06461}.

\bibitem[Wang et~al.(2016)Wang, Schaul, Hessel, Hasselt, Lanctot, and
  Freitas]{wang2016dueling}
Ziyu Wang, Tom Schaul, Matteo Hessel, Hado Hasselt, Marc Lanctot, and Nando
  Freitas.
\newblock Dueling network architectures for deep reinforcement learning.
\newblock In \emph{International conference on machine learning}, pp.\
  1995--2003. PMLR, 2016.

\bibitem[Watkins \& Dayan(1992)Watkins and Dayan]{watkins1992q}
Christopher~JCH Watkins and Peter Dayan.
\newblock Q-learning.
\newblock \emph{Machine learning}, 8\penalty0 (3-4):\penalty0 279--292, 1992.

\bibitem[Ziebart(2010)]{ziebart2010modeling}
Brian~D Ziebart.
\newblock \emph{Modeling purposeful adaptive behavior with the principle of
  maximum causal entropy}.
\newblock Carnegie Mellon University, 2010.

\end{thebibliography}
\bibliographystyle{iclr2022_conference}

\newpage
\appendix

\section{Convergence of Unbiased Soft Q-Learning} \label{proof_of_convergence}

In this section we prove the convergence of each ensemble member $Q^{(k)}$ in ensemble $Q$, to the optimal $Q^*$, with probability 1, under the unbiased soft update rule,
\begin{equation} \label{stochastic_update_rule}
    Q^{(i)} \leftarrow (1 - \alpha_t)Q^{(i)}(s,a) + \alpha_t \bigg( r(s,a) + \gamma \mmax_{a'; \pi_0, w(Q, s')} Q^{(i)}(s', a') \bigg),
\end{equation}
where we define the mellowmax operator with temperature $w$ as
\begin{equation}
    \mmax_{a'; \pi_0, w} Q^{(k)}(s, a) \defeq w \log\E_{a|s\sim\pi_0}\bigg[\exp(Q^{(i)}(s,a)/w)\bigg].
\end{equation}
The temperature $w(Q,s)$ is a function of the ensemble's estimates at state $s$, specifically the value of $w$ that ensures,
\begin{equation} \label{w_match}
    \E_{Q}\bigg[\underset{a; \pi_0, w}{\mmax} \ Q^{(i)}(s, a)\bigg]- \max_a \E_{Q}\bigg[Q^{(i)}(s, a) \bigg].
\end{equation}
We assume a discrete action set with $A$ actions.
%

At a high level, we first show that the temperature-annealed Bellman operator exhibits a contraction property, then use this to
show that for any sequence of temperatures $\{w_t\}$, our ensemble members $\{Q^{(i)}\}$ converge together.  Finally, we show that,
under our temperature selection strategy, this convergence implies that all members converge to $Q^*$.

We first state a few useful properties of the mellowmax function.  Mellowmax is closely related to the temperature-annealed 
$\log \sum \exp$ function (e.g., \citet{liu2011bounding}), from which we can immediately see that
$\mmax_{a;w}$ is convex in $w$, with gradient $\frac{\partial}{\partial w} \mmax_{a;w} Q(a) = H(p_w(a)) + \E_{p_w}[\log \pi_0] = -D(p_w\|\pi_0)$ where
$H(\cdot)$ is the entropy of the annealed Gibbs distribution $p_w(a) \propto \pi_0(a) \exp(Q(a)/w)$ with base measure $\pi_0 >0$ and $D$ is the Kullback-Leibler divergence.
The temperature extremes of $\mmax$ are,
\begin{align*}
\lim_{w\rightarrow 0} \mmax_{a;w} Q(a) &= \max_a Q(a) &
\lim_{w\rightarrow \infty} \mmax_{a;w} Q(a) &= \sum_a \pi_0(a) Q(a).
\end{align*}
Correspondingly, the limiting Gibbs distributions are $p_\infty = \pi_0$ and $p_0$ a distribution that puts uniform weight on the maxima of $Q(a)$.

Finally, we have that
\begin{align*}
w \geq  w' \qquad &\Rightarrow \qquad \mmax_{a,w'} Q(i,a) \geq \mmax_{a,w} Q(a),
\end{align*}
i.e., $\mmax$ is monotonic decreasing in $w$, and that commutation results in a bound:
\begin{align*}
w \geq w' \qquad &\Rightarrow \qquad 
\mmax_{i,w} \mmax_{a,w'} Q(i,a) \geq \mmax_{a,w'} \mmax_{i,w} Q(i,a) 
\end{align*}
with equality if $w=w'$.

Now, define the Bellman operator $\C{B}_{w}$ and denote the optimal fixed point for temperature $w$ as $Q_w^*$, so that:
\begin{equation}
    \C{B}_w[Q_w^*](s, a) = \E[r(s,a)] + \gamma \E_{s' \sim p(\cdot|s,a)}[ \mmax_{a'; \pi_0, w} Q_w^*(s', a') ]
\end{equation}
Then, we can show that,
%
\begin{proof} \label{bellman_contraction}
    Lemma \ref{bellman_contraction_lemma}. Following e.g., \citet{Fox2016Taming}, let
    \begin{equation*}
        \epsilon = \| Q^{(i)} - Q^{(j)}\|_\infty = \max_{s,a} | Q^{(i)}(s,a) - Q^{(j)}(s,a) | 
    \end{equation*}
    Then, for any $s',a'$, 
    \begin{align*}
        \mmax_{a'; \pi_0, w} Q^{(i)}(s', a') &\le \mmax_{a'; \pi_0, w} \bigg( Q^{(j)}(s', a') + \epsilon \bigg) 
        = \epsilon + \mmax_{a'; \pi_0, w} Q^{(j)}(s', a')
    \end{align*}
    and similarly,
    \begin{equation*}
        \mmax_{a'; \pi_0, w} Q^{(i)}(s', a') \ge -\epsilon + \mmax_{a'; \pi_0, w} Q^{(j)}(s', a')
    \end{equation*}
    Therefore
    \begin{equation*}
        \forall s',a' \qquad
        |\mmax_{a'; \pi_0, w} Q^{(i)}(s', a') - \mmax_{a'; \pi_0, w} Q^{(j)}(s', a')| \le \epsilon = \|Q^{(i)}-Q^{(j)}\|_\infty
    \end{equation*}
    Applying this to the definition of the Bellman updates on $Q^{(i)}$ and $Q^{(j)}$ gives,
    \begin{align*}
        \| \C{B}_w[Q^{(i)}] - \C{B}_w[Q^{(j)}] \|_\infty &= \max_{s, a} | \C{B}_w[Q^{(i)}](s, a) - \C{B}_w[Q^{(j)}](s, a) | \\
        &= \gamma \max_{s, a} | \E_{s'}[ \mmax_{a'; \pi_0, w} Q^{(i)}(s', a') - \mmax_{a'; \pi_0, w} Q^{(j)}(s', a') ] | \\ 
        &\le \gamma \| Q^{(i)} - Q^{(j)} \|_\infty
    \end{align*}
\end{proof}

To examine the convergence of our ensemble, we make use of a theorem from \cite{jaakkola1994convergence}:
\begin{theorem}[\cite{jaakkola1994convergence}] \label{thm:jaakkola}
    The random process $\{ \Delta_{t}^{(k,k')}\}$ in form $\Delta_{t+1}^{(k,k')} = (1-\alpha_t(s,a))\Delta_{t}^{(k,k')} + \alpha_t(s,a)F_{t}^{(k,k')}$ taking values in $\R$ converges to 0 w.p.1 if,
    \begin{enumerate}
        \item The number of states $S$ and actions $A$ are finite;
        \item The learning rate obeys $\sum_t \alpha_t(s,a) = \infty$ and $\sum_t \alpha_t^2(s,a) < \infty$;
        \item $\| \E [F^{(k,k')}_t|P_t] \|_W < \gamma \| \Delta_t \|_W$, where $\gamma \in (0,1)$;
        \item $\Var[F^{(k,k')}_t|P_t] \le C(1+\| \Delta_t \|_W)^2$.
    \end{enumerate}
    for some weighted max norm $\|\cdot\|_W$ and where $P_t$ denotes the past history of the algorithm.
\end{theorem}

Now we can show that our ensemble members converge together under a standard stochastic update on each state $s$:

\begin{proof} \label{UQL_convergence_proof}
    Theorem \ref{UQL_convergence}. Our stochastic update has the form,
    \begin{equation}
        Q^{(i)}_{t+1}(s,a) = (1 - \alpha_t(s,a))Q^{(i)}_{t}(s,a) + \alpha_t(s,a)[r_t(s,a) + \gamma \mmax_{a', \pi_0; w_t} Q^{(i)}_{t}(s',a')]
    \end{equation}
    for some sequence of temperatures $w_t$, which may depend on the current ensemble.
    
    For each pair of ensemble members $i,j$, define
    \begin{equation}
        \Delta_{t}^{(i,j)}(s,a) = Q^{(i)}_{t}(s,a) - Q^{(j)}_{t}(s,a).
    \end{equation}
    The stochastic update can be rewritten as
    \begin{multline*}
    \Delta_{t+1}^{(i,j)}(s,a) = (1 - \alpha_t(s,a))\Delta_{t}^{(i,j)}(s,a) \\
        + \alpha_t(s,a)[r_t^{(i)} + \gamma \mmax_{a', \pi_0; w_t} Q^{(i)}_{t}(s^{(i)},a') - r_t^{(j)} - \gamma \mmax_{a', \pi_0; w_t} Q^{(j)}_{t}(s^{(j)},a')],
    \end{multline*}
    suggesting the definition,
    \begin{equation*}
        F^{(i,j)}_{t}(s,a) = \bigg(r_t^{(i)} + \gamma \mmax_{a', \pi_0; w_t} Q^{(i)}_{t}(s^{(i)},a') - r_t^{(j)} - \gamma \mmax_{a', \pi_0; w_t} Q^{(j)}_{t}(s^{(j)},a') \bigg).
    \end{equation*}
    for which it is easy to see that 
    \begin{align*}
        \E_{r^{(i)}_t,r^{(j)}_t,s^{(i)},s^{(j)}}\Big[\ F^{(i,j)}_{t}(s,a) \ \Big] 
        &= \C{B}_{w_t}[Q^{(i)}_{t}](s,a) - \C{B}_{w_t}[Q^{(j)}_{t}](s,a) \\
        & \leq \gamma \, \| Q^{(i)}_t - Q^{(j)}_t \|_\infty \quad = \quad \gamma\, \|\Delta_t^{(i,j)}\|_\infty
    \end{align*}
    where the inequality follows from Lemma~\ref{bellman_contraction_lemma}.
 
    Finally, as in \citet{jaakkola1994convergence}, the required variance condition holds since $\Var[r(s,a)]$  is bounded and $F$ depends at most linearly on the $Q^{(i)}_t$.
    Thus the conditions of Theorem~\ref{thm:jaakkola} are satisfied.
\end{proof}

Finally, we show that as the ensemble members $\{Q^{(i)}\}$ converge together, our temperature $w_t \rightarrow 0$,
and thus all members converge to $Q^*$.
Suppose we have a collection of functions $\{Q^{(i)}(a)\}$ with,
\begin{align*}
\| Q^{(i)} - Q^{(j)} \|_\infty \leq \epsilon \qquad \forall i,j
\end{align*}
We note that the ensemble average can also be denoted as a $\mmax$, since $\frac{1}{|Q|}\sum_i Q^{(i)}(s,a) = {\displaystyle \mmax_{i,\pi_0;\infty} Q^{(i)}(s,a)}$ with uniform $\pi_0$; we abbreviate as $\displaystyle \wmax_{i;\infty}$ for convenience.

Our temperature $w_t$ is thus defined by the value $w^*$ that satisfies,
\begin{align*}
\wmax_{i;\infty} \wmax_{a;w^*} Q^{(i)}(a) = \wmax_{a;0} \wmax_{i,\infty} Q^{(i)}(a).
\end{align*}
This $w^*$ must exist since $\mmax$ is monotone in $w$ and,
\begin{align*}
\wmax_{i;\infty} \wmax_{a;0} Q^{(i)}(a) \geq \wmax_{a;0} \wmax_{i,\infty} Q^{(i)}(a) \geq \wmax_{a;\infty} \wmax_{i;\infty} Q^{(i)}(a) = \wmax_{i;\infty} \wmax_{a;\infty} Q^{(i)}(a);
\end{align*}
the value $w^*$ is unique if the left and right terms are not equal.

If $\|Q^{(i)}-Q^{(j)}\|_\infty \leq \epsilon$, we can easily see that
\begin{align*}
\wmax_{i;\infty} \wmax_{a;0} Q^{(i)} \geq \wmax_{a;0} \wmax_{i;\infty} Q^{(i)} \geq \wmax_{i;\infty} \wmax_{a;0} Q^{(i)}(a)-\epsilon
\end{align*}
since, defining $a^i = \arg\max_a Q^{(i)}(a)$, 
\begin{multline*}
\wmax_{i;\infty}\wmax_{a;0} Q^{(i)} = \wmax_{i;\infty} Q^{(i)}(a^i) \leq \wmax_{i;\infty} \Big[Q^{(1)}(a^i) + \epsilon\Big] \\
  \leq \wmax_{i;\infty} \Big[ Q^{(1)}(a^1)+\epsilon \Big]
  \leq \wmax_{a;0} \wmax_{i;\infty} \Big[Q^{(i)}(a) +2\epsilon\Big]
\end{multline*}
where $Q^{(1)}$ can be any of the functions in the set $\{Q^{(i)}\}$.

Thus we have that our optimal temperature's $\mmax$ value is close to that of $\max$:
\begin{align*}
\wmax_{i;\infty}\wmax_{a;0} Q^{(i)} - \wmax_{i;\infty} \wmax_{a;w^*} Q^{(i)} \quad \leq \quad 2\epsilon
\end{align*}

Now, since $\displaystyle \wmax_{i;\infty}\wmax_{a;w} Q^{(i)}(a)$ is decreasing in $w$, we will have
\begin{align*}
0 &\leq w^* \leq w &&\mbox{if}& \wmax_{i;\infty}\wmax_{a;w} Q^{(i)}(a) &\leq \wmax_{i;\infty}\wmax_{a;0} Q^{(i)}(a)-2\epsilon \leq \wmax_{a;0}\wmax_{i;\infty} Q^{(i)}(a)
\end{align*}
So we need to find a value of $w$ that reduces the $\mmax$ value by at least $2\epsilon$.

Assume that each $Q^{(i)}$ is non-constant, i.e., $\max_a Q^{(i)}(a) - \min_a Q^{(i)}(a) = \delta > 0$.
This ensures that $w^*$ is unique, and (equivalently) that the derivative of $\mmax$ at $w=0$
is non-zero%
\footnote{If $\frac{\partial}{\partial w} \mmax = 0$, then all temperatures $w$ produce the same value,
and we may take $w=0$ without loss of generality.}%
; in particular, if only one action value $Q(a')$ is smaller than the others, 
the slope of $\mmax$ will be $\lim_{w\rightarrow 0} H(p_w) + \E_{p_w}[\log \pi_0] < \max_a \log[1-\pi_0(a)]$ (since $p_w$ will place zero mass on some action $a$).

Thus, as $\epsilon \rightarrow 0$, we have $$w^* \leq \frac{2\epsilon}{-\log[1-\min_a \pi_0(a)]}$$

Therefore, the ensemble members $\{Q^{(i)}\}$ converge to each other, and as they do, $w \rightarrow 0$, so that all
$Q^{(i)}$ are converging to $Q^*$, the optimal $Q$-function under the standard Bellman recursion.

\newpage

\section{Ablation Studies} \label{ablation}
    \begin{figure}[h] 
        \centering
        \includegraphics[width=0.85\textwidth]{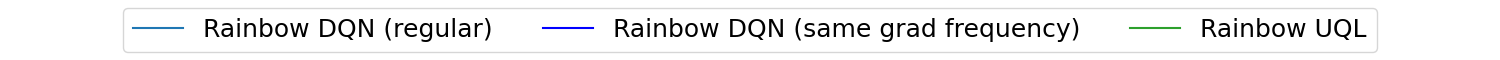} \\
        \includegraphics[width=0.19\textwidth]{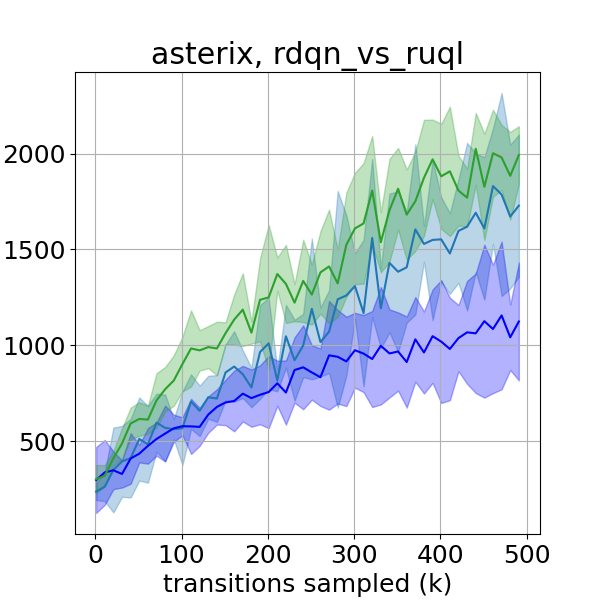}
        \includegraphics[width=0.19\textwidth]{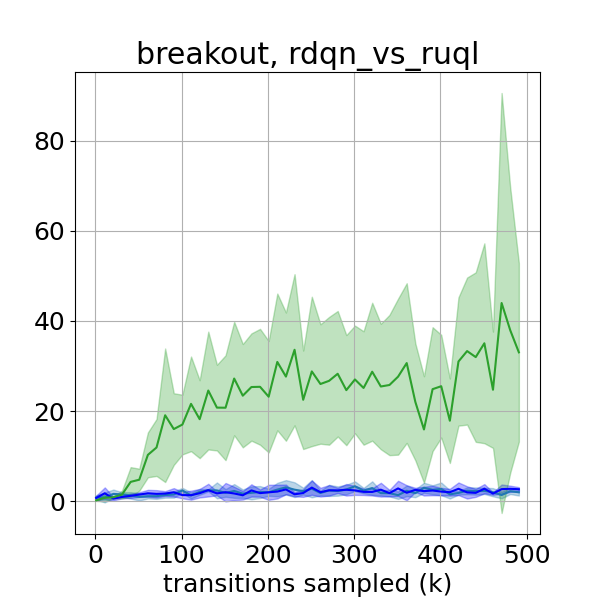}
        \includegraphics[width=0.19\textwidth]{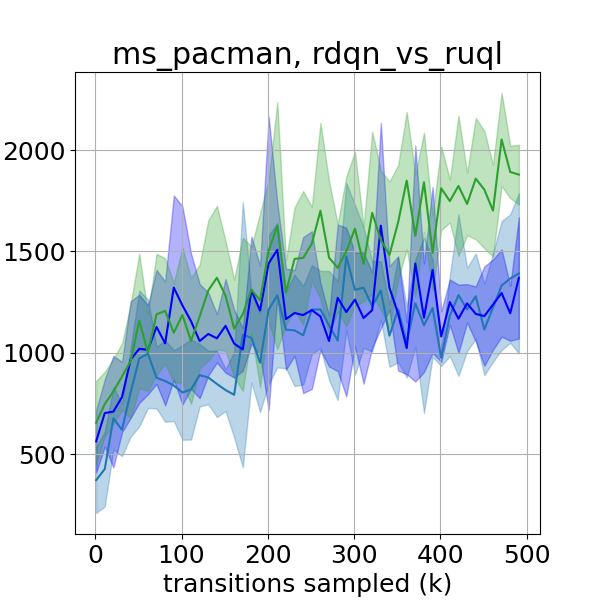}
        \includegraphics[width=0.19\textwidth]{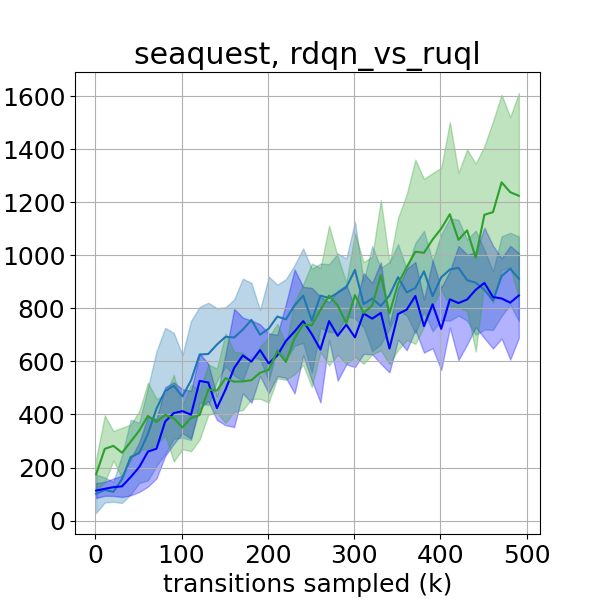}
        \includegraphics[width=0.19\textwidth]{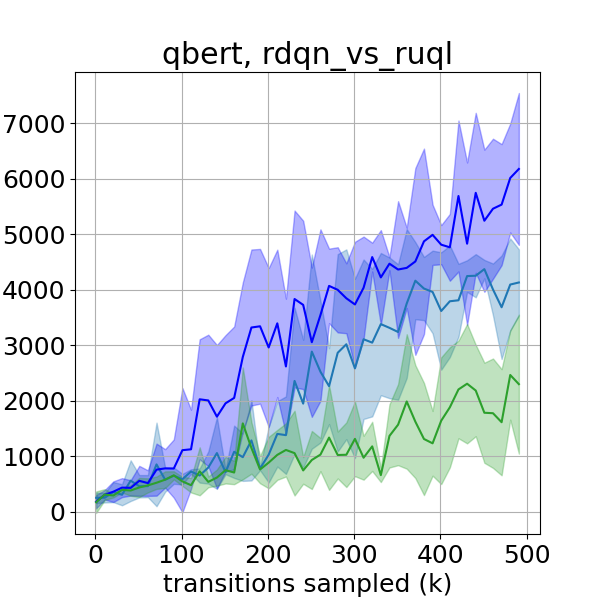}
        \caption{Algorithm (Rainbow, Rainbow with same gradient update frequency as UQL, UQL) average score over 5 runs}
        
    \end{figure}
    \begin{figure}[h] 
        \centering
        \includegraphics[width=0.85\textwidth]{figures/legends/original_legend.png} \\
        \includegraphics[width=0.24\textwidth]{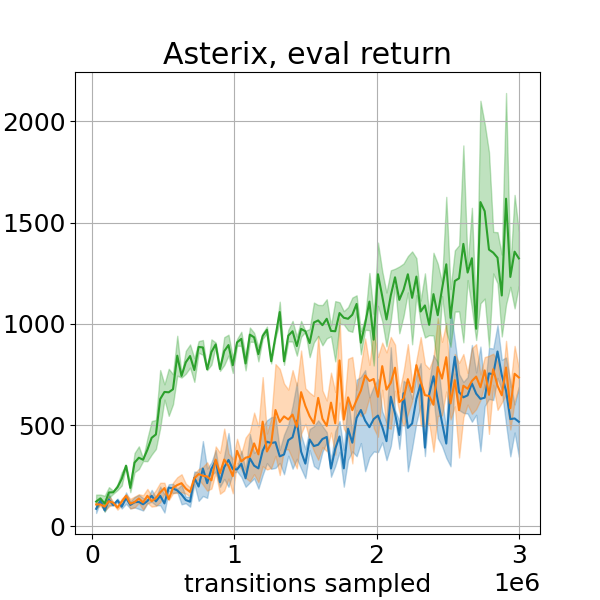}
        \includegraphics[width=0.24\textwidth]{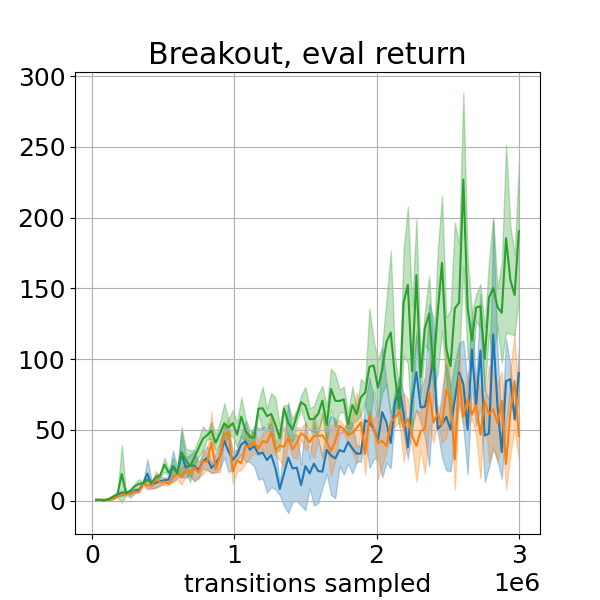}
        \includegraphics[width=0.24\textwidth]{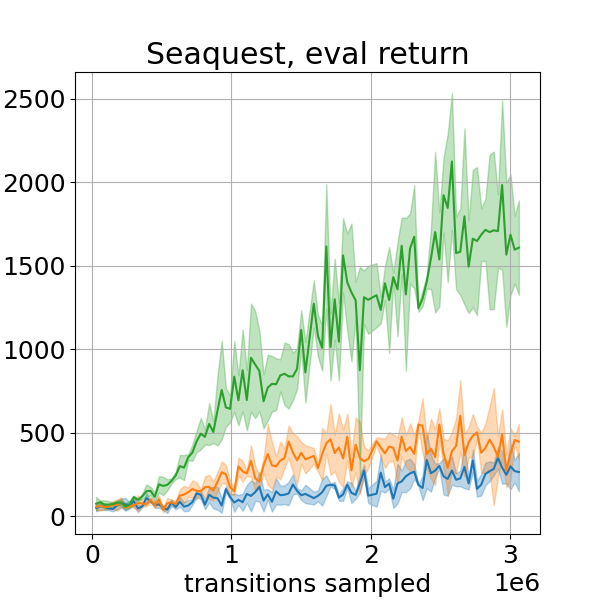}
        \includegraphics[width=0.24\textwidth]{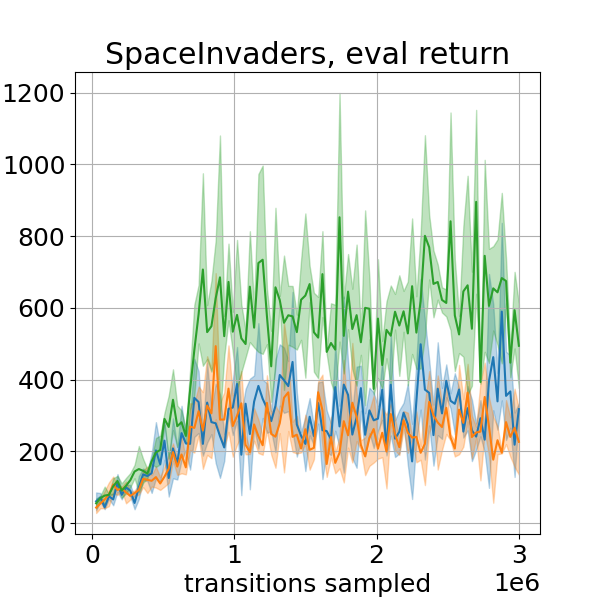} \\
        \includegraphics[width=0.24\textwidth]{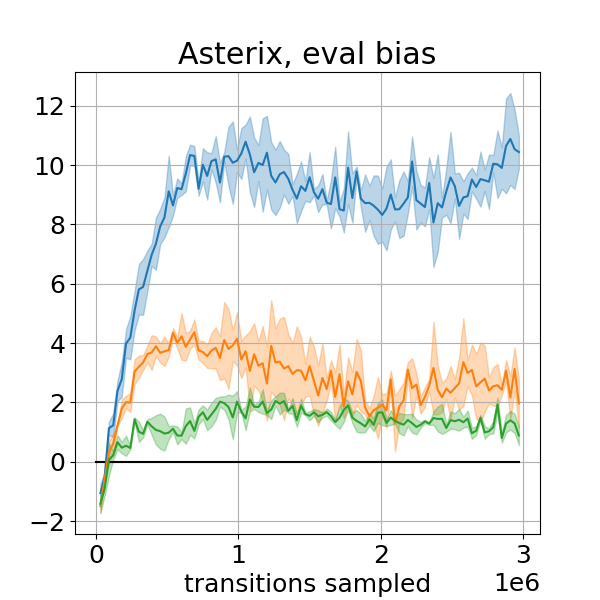}
        \includegraphics[width=0.24\textwidth]{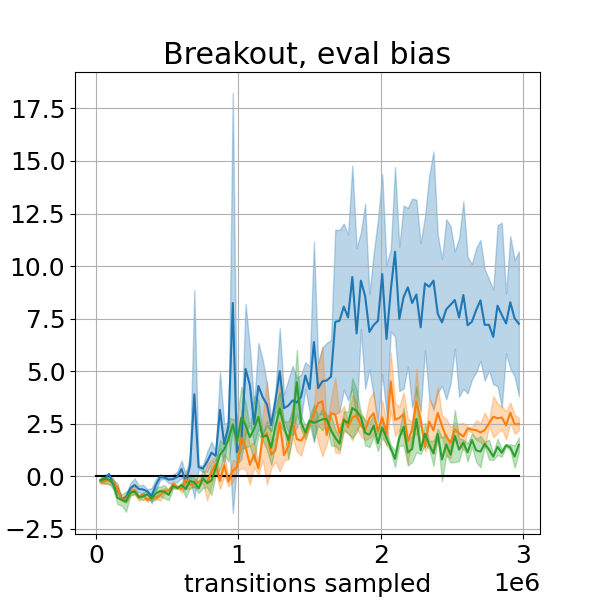}
        \includegraphics[width=0.24\textwidth]{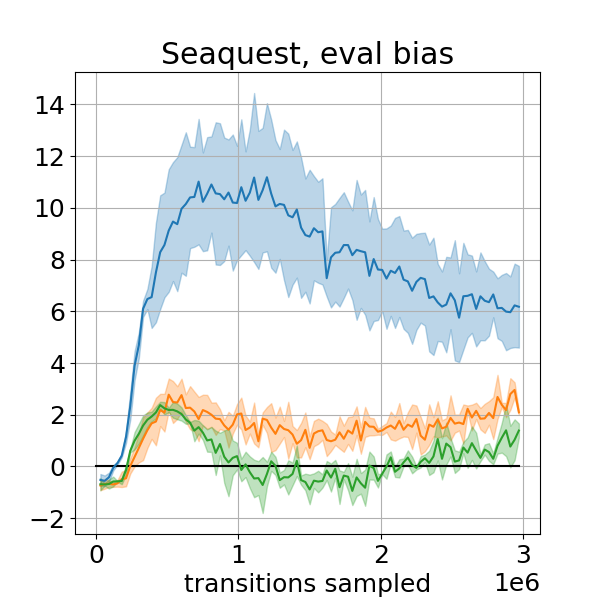}
        \includegraphics[width=0.24\textwidth]{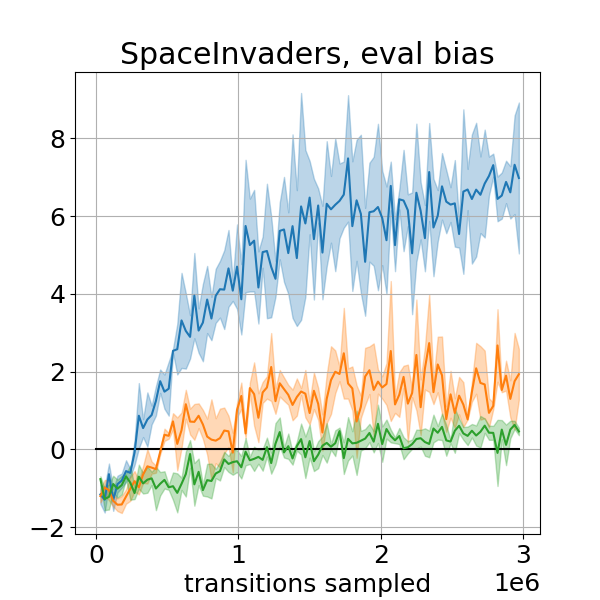}
        
        \caption{First row: Average score over 3 runs. Second row: Initial state discounted value estimation bias averaged over 3 runs. Error bar is the standard deviation of the reported value over 3 runs. Prioritized replay, Dueling Networks are not used in these experiments. The replay buffer is completely shared between algorithms.}
        \label{fig:shared_buffer}
    \end{figure}
    \begin{figure}[h]
            \centering
            \includegraphics[width=0.85\textwidth]{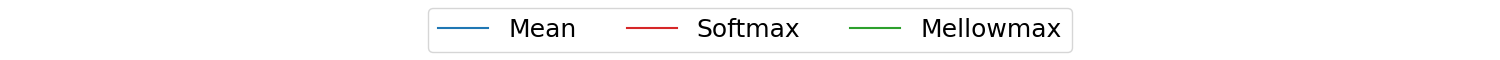} \\
            \includegraphics[width=0.19\textwidth]{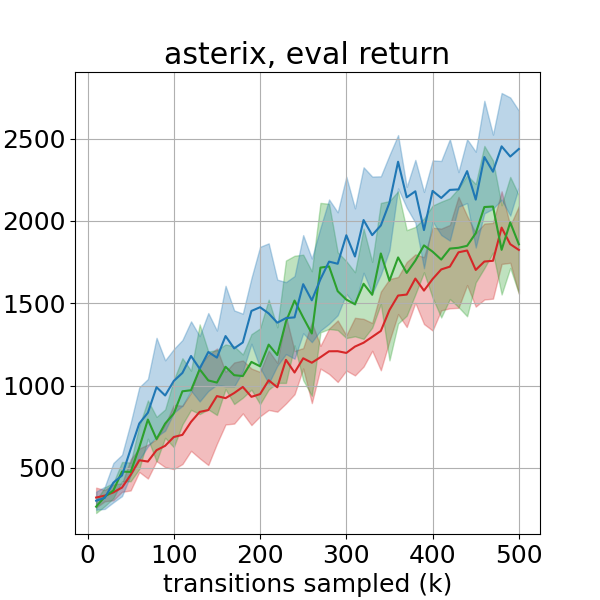}
            \includegraphics[width=0.19\textwidth]{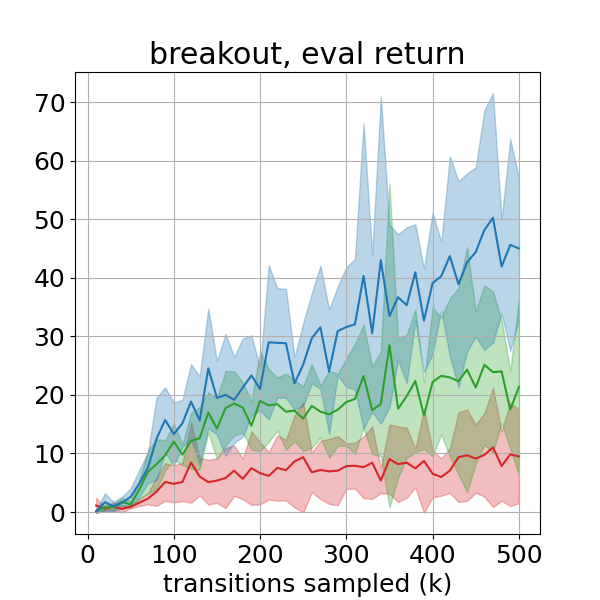}
            \includegraphics[width=0.19\textwidth]{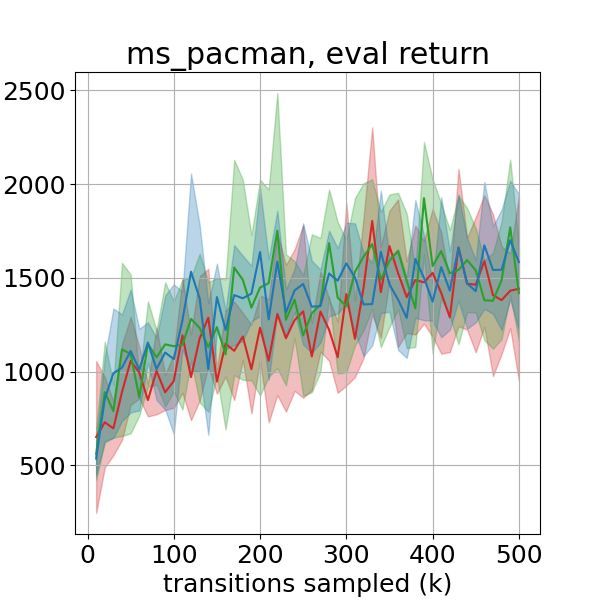}
            \includegraphics[width=0.19\textwidth]{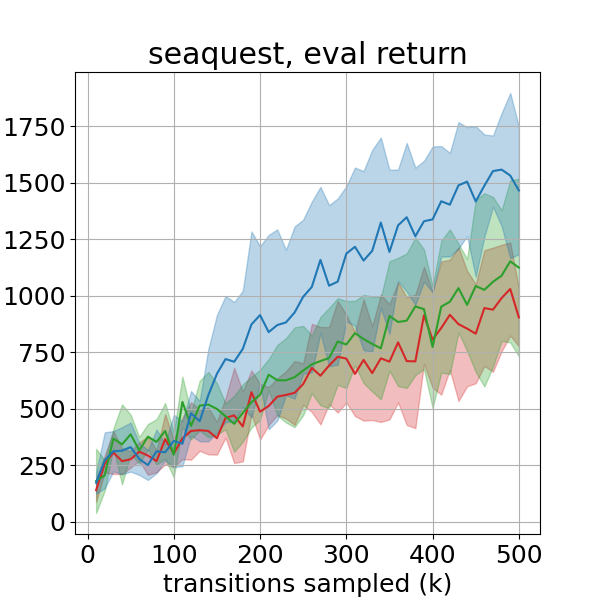}
            \includegraphics[width=0.19\textwidth]{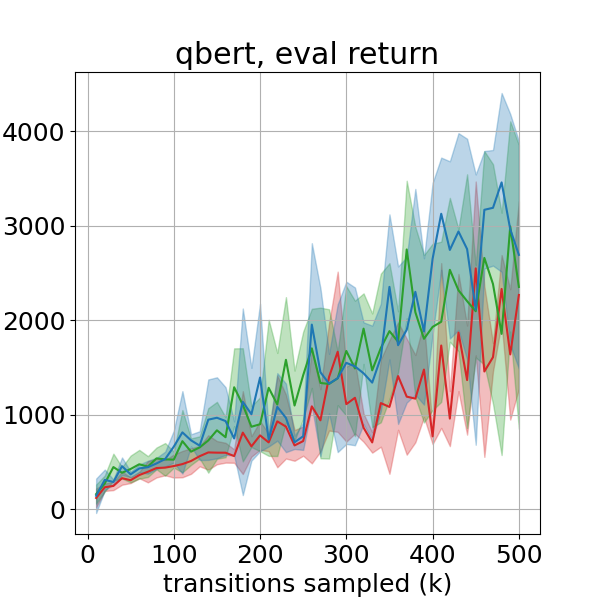} \\
            \includegraphics[width=0.19\textwidth]{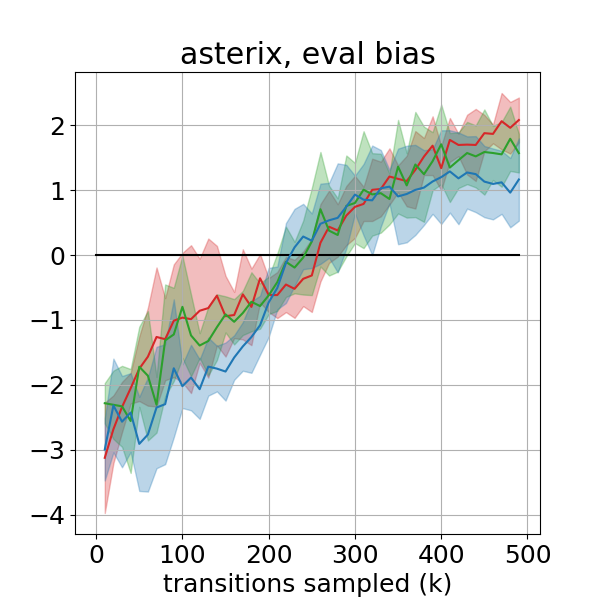}
            \includegraphics[width=0.19\textwidth]{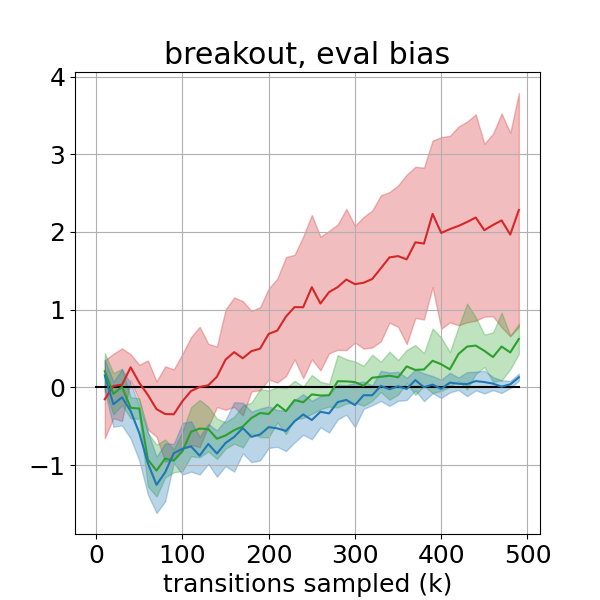}
            \includegraphics[width=0.19\textwidth]{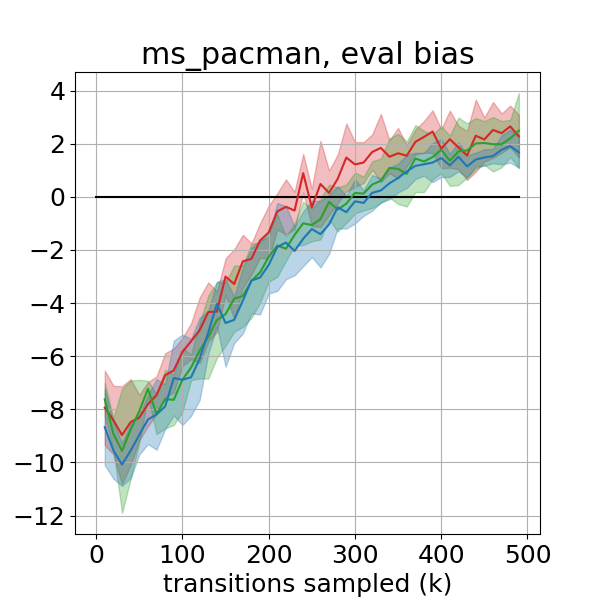}
            \includegraphics[width=0.19\textwidth]{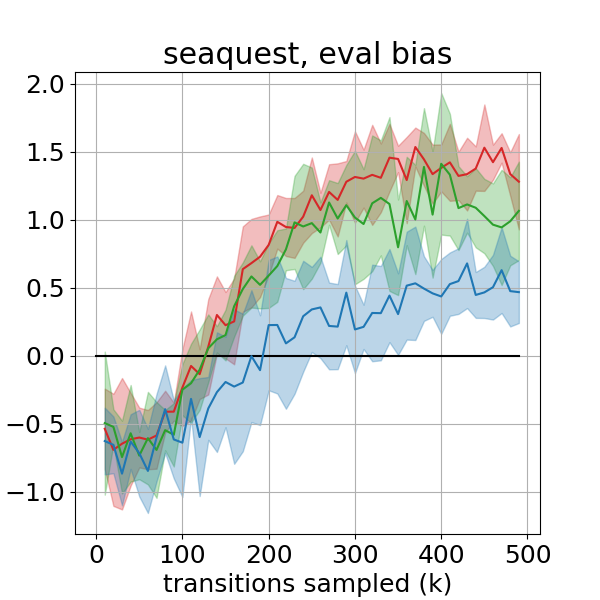}
            \includegraphics[width=0.19\textwidth]{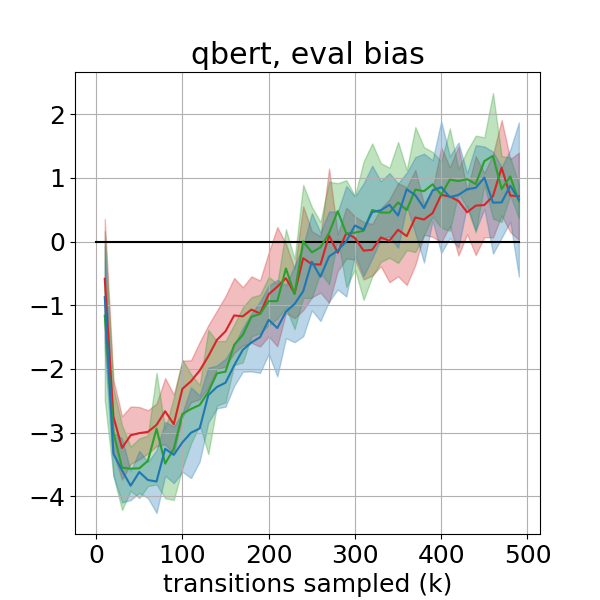}
            \caption{First row: Average score over 5 runs, experience is not shared between algorithms. Error bar is the standard deviation over 5 runs. Second row: Initial state discounted value estimation bias averaged over 5 runs, experience is not shared between algorithms. Error bar is the standard deviation over 5 runs.}
            \label{fig:uql_diff_operator}
        \end{figure}

\newpage
\section{Hyperparameter Configurations}
All results in Table \ref{main_results} is generated using the hyperparameter configuration from \cite{DBLP:journals/corr/abs-2007-04938}, except hyperparameters for $\kappa$ and searching $\beta$. No tuning is done on the hyperparameters. $\rightarrow$ denotes linear schedule.
\begin{table}[H]
    \caption{Hyperparameter configuration of UQL used in experiments in Atari domain}
    \label{hyperparameters}
    \begin{center}
    \begin{tabular}{ll}
    \multicolumn{1}{c}{\bf Hyperparameter} 
        &\multicolumn{1}{c}{\bf Value}
    \\ \hline \\  
    learning rate & $1\times10^{-4}$ \\ 
    optimization algorithm & Adam \\
    mini-batch size & 32 \\
    dueling network & True \\
    hidden size of A and V head & 256 \\
    number of interactions per target network update & 2000 \\
    number of interactions per gradient update & 2 \\
    discount factor & 0.99 \\
    ensemble size & 5 \\
    $\beta$ search range & $[1 \times 10^{-20}, 2 \times 10^6]$ \\
    $\beta$ search max iteration & 35 \\
    correction constant $\kappa$ & 1\\
    number of interactions before learning start & 1600 \\
    UCB exploration $\lambda$ & 1 \\
    replay buffer capacity & $5 \times 10^5$ \\
    priority weight & $0.5 \rightarrow 0$ \\
    importance sampling exponent & $0.4 \rightarrow 1$ \\

    \end{tabular}
    \end{center}
    \end{table}

\newpage
\section{Performance Normalized by Human Score}
    Performance at 500k interactions. The results of UQL ($\kappa=1$ and $\kappa=0.5$) show the average score and standard deviation of 5 runs. For Random, we report the numbers reported in \citet{DBLP:journals/corr/abs-2007-04938}. For PPO, Rainbow, we report the number reported in \citet{DBLP:journals/corr/abs-1903-00374}. 
\begin{figure}[H] 
    \centering
    \includegraphics[width=0.8\textwidth]{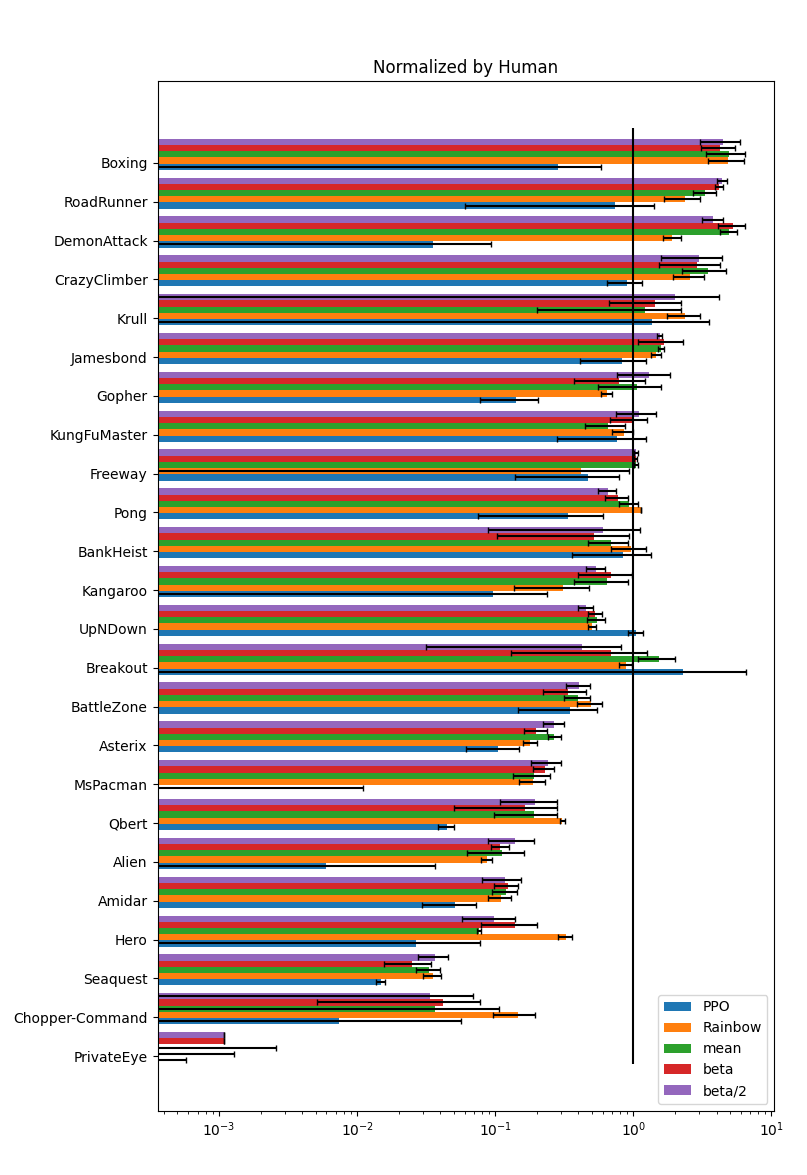}
    \caption{Performance is computed using: $\frac{\text{Score}_{\text{Algorithm}} - \text{Score}_{\text{Random}}}{\text{Score}_{\text{Human}} - \text{Score}_{\text{Random}}}$}
\end{figure}

\begin{landscape}
\section{Performance Compare to Human at 500k interaction} \label{main_results}
\begin{table}[H] 
\begin{center}
\begin{tabular}{llllllllllllll}
\multicolumn{1}{l}{Environment}  
    &\multicolumn{1}{l}{Random} 
    &\multicolumn{1}{l}{PPO} 
    &\multicolumn{1}{l}{}
    &\multicolumn{1}{l}{Rainbow}
    &\multicolumn{1}{l}{}
    &\multicolumn{1}{l}{Mean}
    &\multicolumn{1}{l}{}
    &\multicolumn{1}{l}{$\kappa=1$}
    &\multicolumn{1}{l}{}
    &\multicolumn{1}{l}{$\kappa=0.5$}
    &\multicolumn{1}{l}{}
    &\multicolumn{1}{l}{Human}
\\ \hline \\  
Alien           & 227.8   & 269.0            & (203.4)   & 828.6            & (54.2)   & 997.00            & (326.51)  & 979.17           & (108.00)   & \textbf{1189.60}  & (339.96)   & 7128.0      \\
Amidar          & 5.8     & 93.2             & (36.7)    & 194.0            & (34.9)   & 210.40            & (41.61)   & \textbf{216.25}  & (41.70)    & 207.25            & (63.00)    & 1720.0      \\
Asterix         & 210.0   & 1085.0           & (354.8)   & 1702.7           & (162.8)  & \textbf{2437.00}  & (232.37)  & 1858.00          & (298.69)   & 2420.50           & (376.80)   & 8503.0      \\
BankHeist       & 14.2    & 641.0            & (352.8)   & \textbf{727.3}   & (198.3)  & 523.90            & (161.51)  & 395.20           & (298.45)   & 459.0             & (371.68)   & 753.0       \\
BattleZone      & 2360.0  & 14400.0          & (6476.1)  & \textbf{19507.1} & (3193.3) & 16280.00          & (2675.00) & 14160.00         & (3777.35)  & 16460.0           & (2510.06)  & 37188.0     \\
Boxing          & 0.1     & 3.5              & (3.5)     & 58.2             & (16.5)   & \textbf{58.26}    & (17.8)    & 50.92            & (14.22)    & 53.21             & (17.10)    & 12.0        \\
Breakout        & 1.7     & \textbf{66.1}    & (114.3)   & 26.7             & (2.4)    & 45.02             & (12.11)   & 21.34            & (14.99)    & 13.67             & (10.41)    & 30.0        \\
Chopper-Command & 811.0   & 860.0            & (285.3)   & \textbf{1765.2}  & (280.7)  & 1052.00           & (404.04)  & 1084.89          & (210.72)   & 1031.0            & (206.29)   & 7388.0      \\
CrazyClimber    & 10780.5 & 33420.0          & (3628.3)  & 75655.1          & (9439.6) & \textbf{97302.00} & (17027.8) & 83237.00         & (19315.91) & 85315.0           & (19750.28) & 35829.0     \\
DemonAttack     & 152.1   & 216.5            & (96.2)    & 3642.1           & (478.2)  & 9150.8            & (1128.25) & \textbf{9737.50} & (1911.48)  & 7039.55           & (1087.00)  & 1971.0      \\
Freeway         & 0.0     & 14.0             & (9.8)     & 12.6             & (15.4)   & 31.25             & (0.95)    & 31.19            & (0.73)     & \textbf{31.53}    & (1.10)     & 30.0        \\
Frostbite       & 65.2    & 214.0            & (10.2)    & \textbf{1386.1}  & (321.7)  & 687.80            & (823.21)  & {895.70}         & (725.57)   & 1214.90           & (955.95)   & -           \\
Gopher          & 257.6   & 560.0            & (118.8)   & 1640.5           & (105.6)  & 2563.8            & (977.46)  & 1965.60          & (792.87)   & \textbf{3063.0}   & (1018.69)  & 2412.0      \\
Hero            & 1027.0  & 1824.0           & (1461.2)  & \textbf{10664.3} & (1060.5) & 3288.42           & (75.7)    & 5161.45          & (1737.55)  & 3941.30           & (1173.67)  & 30826.0     \\
Jamesbond       & 29.0    & 255.0            & (101.7)   & 429.7            & (27.9)   & 465.00            & (21.39)   & \textbf{488.5}   & (147.09)   & 455.5             & (15.44)    & 303.0       \\
Kangaroo        & 52.0    & 340.0            & (407.9)   & 970.9            & (501.9)  & 1972.00           & (798.16)  & \textbf{2112.00} & (858.03)   & 1662.0            & (254.98)   & 3035.0      \\
Krull           & 1598.0  & 3056.1           & (1155.5)  & \textbf{4139.4}  & (336.2)  & 2895.89           & (538.43)  & 3136.31          & (412.84)   & 3749.13           & (1147.11)  & 2666.0      \\
KungFuMaster    & 258.5   & 17370.0          & (10707.6) & 19346.1          & (3274.4) & 15020.00          & (4741.49) & 22109.33         & (6481.43)  & \textbf{24942.0}  & (7754.50)  & 22736.0     \\
MsPacman        & 307.3   & 306.0            & (70.2)    & 1558.0           & (248.9)  & 1583.70           & (362.53)  & 1818.8           & (255.52)   & \textbf{1911.6}   & (372.95)   & 6952.0      \\
Pong            & -20.7   & -8.6             & (14.9)    & \textbf{19.9}    & (0.4)    & 12.56             & (8.30)    & 6.73             & (8.38)     & 2.61              & (5.56)     & 15.0        \\
PrivateEye      & 24.9    & 20.0             & (40.0)    & -6.2             & (89.8)   & -51.68            & (182.02)  & \textbf{100.0}   & (0.00)     & \textbf{100.0}    & (0.00)     & 69571.0     \\
Qbert           & 163.9   & 757.5            & (78.9)    & \textbf{4241.7}  & (193.1)  & 2691.00           & (1203.76) & 2352.00          & (1501.12)  & 2751.25           & (1123.12)  & 13455.0     \\
RoadRunner      & 11.5    & 5750.0           & (5259.9)  & 18415.4          & (5280.0) & 25970.00          & (4858.74) & 32766.66         & (2299.01)  & \textbf{34581.00} & (2996.29)  & 7845.0      \\
Seaquest        & 68.4    & 692.0            & (48.3)    & 1558.7           & (221.2)  & 1466.20           & (283.03)  & 1124.6           & (393.02)   & \textbf{1597.80}  & (369.56)   & 42055.0     \\
UpNDown         & 533.4   & \textbf{12126.0} & (1389.5)  & 6120.7           & (356.8)  & 6608.25           & (888.00)  & 6416.66          & (646.50)   & 5575.50           & (578.81)   & 11693.0     \\
\end{tabular}
\end{center}
\caption{Performance at 500k interactions. The results of UQL ($\kappa=1$ and $\kappa=0.5$) show the average score and standard deviation of 5 runs. For Random, we report the numbers reported in \citet{DBLP:journals/corr/abs-2007-04938}. For PPO, Rainbow, we report the number reported in \citet{DBLP:journals/corr/abs-1903-00374}. Best performance is shown in bold.}
\end{table}
\end{landscape}

\begin{landscape}
\section{Performance Compare to Human at 100k interaction}
\begin{table}[H] 
\fontsize{6}{8}\selectfont 
\begin{center}
\begin{tabular}{lllllllllllllllllll}
\multicolumn{1}{l}{Environment}  
    &\multicolumn{1}{l}{Random} 
    &\multicolumn{1}{l}{PPO} 
    &\multicolumn{1}{l}{}
    &\multicolumn{1}{l}{Rainbow}
    &\multicolumn{1}{l}{}
    &\multicolumn{1}{l}{SUNRISE}
    &\multicolumn{1}{l}{SimPLe}
    &\multicolumn{1}{l}{}
    &\multicolumn{1}{l}{CURL}
    &\multicolumn{1}{l}{DrQ}
    &\multicolumn{1}{l}{Mean}
    &\multicolumn{1}{l}{}
    &\multicolumn{1}{l}{$\kappa=1$}
    &\multicolumn{1}{l}{}
    &\multicolumn{1}{l}{$\kappa=0.5$}
    &\multicolumn{1}{l}{}
    &\multicolumn{1}{l}{Human}
\\ \hline \\  
Alien          & 184.8  & 291.0    &(40.3)& 290.6   &(14.8)&\textbf{872.0}& 616.9   &(252.2)& 558.2   & 761.4   &   625.20 &(40.5)&   647.70 &(45.35)&   590.20 &(112.66)& 7128.0 \\
Amidar         & 11.8   & 56.5     &(20.8)& 20.8    &(2.3)& 122.6   & 74.3    &(28.3)&\textbf{142.1}& 97.3    &    93.59 &(21.52)&    81.05 &(16.44)&   101.56 &(33.94)& 1720.0 \\
Asterix        & 248.8  & 385.0    &(104.4)& 285.7   &(9.3)& 755.0   &\textbf{1128.3}&(211.8)& 734.5   & 637.5   &  1028.50 &(193.14)&   830.50 &(204.04)&   949.50 &(151.59)& 8503.0 \\
BankHeist      & 15.0   & 16.0     &(12.4)& 34.5    &(2.0)&\textbf{266.7}& 34.2    &(29.2)& 131.6   & 196.6   &   262.90 &(64.86)&   199.90 &(76.4)&   233.80 &(23.98)& 753.0  \\
BattleZone     & 2895.0 & 5300.0   &(3655.1)& 3363.5  &(523.8)&\textbf{15700.0}& 4031.2  &(1156.1)& 14870.0 & 13520.6 & 10640.00 &(3612.12)&  9260.00 &(2776.94)& 12800.00 &(2774.35)& 37188.0\\
Boxing         & 0.3    & -3.9     &(6.4)& 0.9     &(1.7)& 6.7     &\textbf{7.8}&(10.1)& 1.2     & 6.9     &     5.96 &(4.33)&     2.71 &(3.6)&     7.49 &(3.1)& 12.0   \\
Breakout       & 0.9    & 5.9      &(3.3)& 3.3     &(0.1)& 1.8     &\textbf{16.4}&(6.2)& 4.9     & 14.5    &    13.32 &(5.37)&    12.04 &(3.12)&     7.83 &(6.55)& 30.0   \\
ChopperCommand & 671.0  & 730.0    &(199.0)& 776.6   &(59.0)& 1040.0  & 979.4   &(172.7)&\textbf{1058.5}& 646.6   &   840.00 &(146.9)&   892.00 &(89.92)&   882.00 &(197.78)& 7388.0 \\
CrazyClimber   & 7339.5 & 18400.0  &(5275.1)& 12558.3 &(674.6)& 22230.0 &\textbf{62583.6}&(16856.8)& 12146.5 & 19694.1 & 60035.00 &(26988.73)& 52798.00 &(22817.56)& 44838.00 &(15690.94)& 35829.0\\
DemonAttack    & 140.0  & 192.5    &(83.1)& 431.6   &(79.5)& 919.8   & 208.1   &(56.8)& 817.6   &\textbf{1222.2}&   741.05 &(232.55)&   522.55 &(109.47)&   450.65 &(102.49)& 1971.0 \\
Freeway        & 0.0    & 8.0      &(9.8)& 0.1     &(0.1)&\textbf{30.2}& 16.7    &(15.7)& 26.7    & 15.4    &    14.24 &(1.99)&    10.67 &(6.85)&     5.18 &(8.45)& 30.0   \\
Frostbite      & 74.0   & 214.0    &(10.2)& 1386.1  &(321.7)&\textbf{2026.7}& 65.2    &(31.5)& 1181.3  & 449.7   &   271.30 &(56.41)&   239.50 &(20.21)&   394.60 &(278.98)& 0.0\\
Gopher         & 245.9  & 246.0    &(103.3)& 748.3   &(105.4)& 654.7   & 596.8   &(183.5)& 669.3   & 598.4   &\textbf{896.4}&(194.44)&   882.60 &(241.3)&   890.20 &(188.83)& 2412.0 \\
Hero           & 224.6  & 569.0    &(1100.9)& 2676.3  &(93.7)&\textbf{8072.5}& 2656.6  &(483.1)& 6279.3  & 4001.6  &  1277.35 &(1433.68)&  2969.10 &(168.43)&  2442.00 &(1222.07)& 30826.0\\
Jamesbond      & 29.2   & 65.0     &(46.4)& 61.7    &(8.8)& 390.0   & 100.5   &(36.8)&\textbf{471.0}& 272.3   &   217.50 &(71.15)&   203.00 &(120.58)&    99.00 &(27.18)& 303.0  \\
Kangaroo       & 42.0   & 140.0    &(102.0)& 38.7    &(9.3)&\textbf{2000.0}& 51.2    &(17.8)& 872.5   & 1052.4  &   164.00 &(143.47)&   456.00 &(291.73)&   250.00 &(140.43)& 3035.0 \\
Krull          & 1543.3 & 3750.4   &(3071.9)& 2978.8  &(148.4)& 3087.2  & 2204.8  &(776.5)&\textbf{4229.6}& 4002.3  &  2563.76 &(124.32)&  2163.85 &(297.58)&  2407.25 &(407.11)& 2666.0 \\
KungFuMaster   & 616.5  & 4820.0   &(983.2)& 1019.4  &(149.6)& 10306.7 & 14862.5 &(4031.6)& 14307.8 & 7106.4  & 13541.00 &(2844.02)&\textbf{20094.0}&(3741.89)& 19685.00 &(8715.99)& 22736.0\\
MsPacman       & 235.2  & 496.0    &(379.8)& 364.3   &(20.4)&\textbf{1482.3}& 1480.0  &(288.2)& 1465.5  & 1065.6  &  1066.90 &(399.22)&  1135.50 &(239.87)&  1157.50 &(323.51)& 6952.0 \\
Pong           & -20.4  & -20.5    &(0.6)& -19.5   &(0.2)& -19.3   &\textbf{12.8}&(17.2)& -16.5   & -11.4   &   -14.05 &(2.62)&   -16.15 &(5.48)&   -18.52 &(2.82)& 15.0   \\
PrivateEye     & 26.6   & 10.0     &(20.0)& 42.1    &(53.8)& 100.0   & 35.0    &(60.2)&\textbf{218.4}& 49.2    &   -35.49 &(270.98)&   100.00 &(0.0)&    77.00 &(34.87)& 69571.0\\
Qbert          & 166.1  & 362.5    &(117.8)& 235.6   &(12.9)&\textbf{1830.8}& 1288.8  &(1677.9)& 1042.4  & 1100.9  &   662.50 &(168.19)&   525.00 &(84.62)&   470.25 &(123.55)& 13455.0\\
RoadRunner     & 0.0    & 1430.0   &(760.0)& 524.1   &(147.5)& 11913.3 & 5640.6  &(3936.6)& 5661.0  & 8069.8  & 16366.00 &(7868.77)& 16116.00 &(7262.98)&\textbf{18448.0}&(8074.48)& 7845.0 \\
Seaquest       & 61.1   & 370.0    &(103.3)& 206.3   &(17.1)& 570.7   &\textbf{683.3}&(171.2)& 384.5   & 321.8   &   357.60 &(116.15)&   296.00 &(100.02)&   333.00 &(91.47)& 42055.0\\
UpNDown        & 488.4  & 2874.0   &(1105.8)& 1346.3  &(95.1)&\textbf{5074.0}& 3350.3  &(3540.0)& 2955.2  & 3924.9  &  2572.10 &(920.03)&  3191.10 &(484.26)&  2283.30 &(434.12)& 11693.0\\
\end{tabular}
\end{center}
\caption{Performance at 100k interactions. The results of UQL ($\kappa=1$ and $\kappa=0.5$) show the average score and standard deviation of 5 runs. For Random and SUNRISE, we report the numbers reported in \citet{DBLP:journals/corr/abs-2007-04938}. For PPO, Rainbow, SimPLe, CURL, and DrQ, we report the number reported in \citet{DBLP:journals/corr/abs-1903-00374}. Best performance is shown in bold.}\label{main_results_100k}
\end{table}
\end{landscape}

\end{document}